\pdfoutput=1

\documentclass{article}

\usepackage{microtype}
\usepackage{graphicx}
\usepackage{subfigure}
\usepackage{booktabs} 
\usepackage{amssymb}
\usepackage{amsmath}
\usepackage{amsthm}
\usepackage[dvipsnames]{xcolor}
\usepackage{caption}

\usepackage{hyperref}

\newtheorem{definition}{Definition}[section]

\newtheorem{proposition}{Proposition}[section]

\usepackage[accepted]{icml2020}

\icmltitlerunning{Efficient Continuous Pareto Exploration in Multi-Task Learning}

\begin{document}

\twocolumn[
\icmltitle{Efficient Continuous Pareto Exploration in Multi-Task Learning}



\icmlsetsymbol{equal}{*}

\begin{icmlauthorlist}
\icmlauthor{Pingchuan Ma}{equal,mit}
\icmlauthor{Tao Du}{equal,mit}
\icmlauthor{Wojciech Matusik}{mit}
\end{icmlauthorlist}

\icmlaffiliation{mit}{MIT CSAIL}

\icmlcorrespondingauthor{Pingchuan Ma}{pcma@csail.mit.edu}

\icmlkeywords{Multi-Objective Optimization, Multi-Task Learning, Hessian-Free Methods}

\vskip 0.3in
]



\printAffiliationsAndNotice{\icmlEqualContribution} 

\newcommand{\m}[1]{\boldsymbol{#1}}

\begin{abstract}
Tasks in multi-task learning often correlate, conflict, or even compete with each other. As a result, a single solution that is optimal for all tasks rarely exists. Recent papers introduced the concept of Pareto optimality to this field and directly cast multi-task learning as multi-objective optimization problems, but solutions returned by existing methods are typically finite, sparse, and discrete. We present a novel, efficient method that generates locally continuous Pareto sets and Pareto fronts, which opens up the possibility of continuous analysis of Pareto optimal solutions in machine learning problems. We scale up theoretical results in multi-objective optimization to modern machine learning problems by proposing a sample-based sparse linear system, for which standard Hessian-free solvers in machine learning can be applied. We compare our method to the state-of-the-art algorithms and demonstrate its usage of analyzing local Pareto sets on various multi-task classification and regression problems. The experimental results confirm that our algorithm reveals the primary directions in local Pareto sets for trade-off balancing, finds more solutions with different trade-offs efficiently, and scales well to tasks with millions of parameters.
\end{abstract}

\section{Introduction}
Conflicting objectives are common in machine learning problems: designing a machine learning model takes into account model complexity and generalizability, training a model minimizes bias and variance errors from datasets, and evaluating a model typically involves multiple metrics that are, more often than not, competing with each other. Such trade-offs among objectives often invalidate the existence of one single solution optimal for all objectives. Instead, they give rise to a set of solutions, known as the Pareto set, with varying preferences on different objectives.

In this paper, we are interested in the topic of recovering Pareto sets in deep multi-task learning (MTL) problems. Despite that MTL is inherently a multi-objective problem and trade-offs are frequently observed in theory and practice, most of prior work focused on obtaining one optimal solution that is universally used for all tasks. To solve this problem, prior approaches proposed new model architectures \cite{misra16} or developed new optimization algorithms \cite{kendall18,sener18}. Work on exploring a diverse set of solutions with trade-offs is surprisingly rare and limited to finite and discrete solutions \cite{lin19}. In this work, we address this challenging problem by proposing an efficient method that reconstructs a first-order accurate continuous approximation to Pareto sets in MTL problems.

The significant leap from finding a discrete Pareto set to discovering a continuous one requires a fundamentally novel algorithm. Typically, generating one solution in a Pareto set is a time-consuming process that requires expensive optimization (e.g., training a neural network). In order to obtain an efficient algorithm for computing a continuous Pareto set, it is necessary to exploit local information. Our technical method is inspired by second-order methods in multi-objective optimization (MOO) \cite{hillermeier01b,martin18,schulz18} which connect the local tangent plane, the gradient information, and the Hessian matrices at a Pareto optimal solution all in one concise linear equation. This theorem allows us to construct a continuous, first-order approximation of the local Pareto set. However, naively applying this method to deep MTL scales poorly with the number of parameters (e.g., the number of weights in a neural network) due to its need to compute full Hessian matrices. Motivated by other second-order methods in deep learning \cite{martens10,vinyals12}, we propose to resolve the scalability issue by using Krylov subspace iteration methods, a family of matrix-free, iterative linear solvers, and present a complete algorithm for generating families of continuous Pareto sets in deep MTL.

We empirically evaluate our method on five datasets with various size and model complexity, ranging from MultiMNIST \cite{sabour17} that consists of 60k images and requires a network classifier with only 20k parameters, to UTKFace \cite{zhang17}, an image dataset with 3 objectives and a modern network structure with millions of parameters. The code and data are available online\footnote{https://github.com/mit-gfx/ContinuousParetoMTL}. Experimental results demonstrate that our method generates much denser Pareto sets and Pareto fronts than previous work with small computational overhead compared to the whole MTL training process. We also show in the experiments the continuous Pareto sets can be reparametrized into a low dimensional parameter space, allowing for intuitive manipulation and traversal in the Pareto set. We believe that our efficient and scalable algorithm can open up new possibilities in MTL and foster a deeper understanding of trade-offs between tasks.
\section{Related work}
Multi-task learning (MTL) is a learning paradigm that jointly optimizes a set of tasks with shared parameters. It is generally assumed that information across different tasks can reinforce the training of shared parameters and improve the overall performance in all tasks. However, since MTL problems share some parameters, performances on different tasks compete with each other. Therefore, trade-offs between performances on different tasks are usually prevalent in MTL. A standard strategy to deal with these trade-offs is to formulate a single-objective optimization problem which assigns weights to each task \cite{kokkinos17}. Choosing weights for each task is typically empirical, problem-specific, and tedious. To simplify the process of selecting weights, prior work suggests some heuristics on adaptive weights \cite{chen17,kendall18}. However, this family of methods aims to find one optimal solution for all tasks and is not designed for exploring trade-offs.

Instead of solving a weighted sum of tasks as a single objective, some recent papers directly cast MTL as a multi-objective optimization (MOO) problem and introduce multiple gradient-based methods (MGDA) \cite{fliege00,desideri12,fliege16} to MTL. Sener and Koltun \yrcite{sener18} formally formulate MTL as an MOO problem and propose to use MGDA for training a single optimal solution for all objectives. Another recent approach \cite{lin19}, which is the most relevant to our setting, pushes the frontier further by pointing out the necessity of exploring Pareto fronts in MTL and presents an MGDA-based method to generate a discrete set of solutions evenly distributed on the Pareto front. Each solution in their method requires full training from an initial network, which limits its ability to generate a dense set of Pareto optimal solutions. 

All the methods discussed so far are based on first-order algorithms in MOO and generate either one solution or a finite set of sparse solutions with trade-offs. A clear distinction between our paper and previous work is that we propose replacing discrete solutions with continuous solution families, allowing for a much denser set of solutions and continuous analysis on them. The advance from discrete to continuous solutions requires a second-order analysis tool in MOO \cite{hillermeier01b,martin18,schulz18}, which embeds tangent planes, gradients, and Hessians in one concise linear system. Our work is also related to Hessian-free methods in machine learning \cite{martens10,vinyals12} which rely heavily on Hessian-vector products in neural networks \cite{pearlmutter94} to solve Hessian systems efficiently.
\section{Preliminaries}

In this work, we consider an unconstrained multi-objective optimization problem described by $\m{f}(\m{x}): \mathbb{R}^n\rightarrow\mathbb{R}^m$ where each $f_i(\m{x}):\mathbb{R}^n\rightarrow\mathbb{R},i=1,2,\cdots,m$ represents the objective function of the $i$-th task to be minimized. For any $\m{x},\m{y}\in\mathbb{R}^n$, $\m{x}$ dominates $\mathbf{y}$ if and only if $\m{f}(\m{x})\leq\m{f}(\m{y})$ and $\m{f}(\m{x})\neq\m{f}(\m{y})$. A point $\m{x}$ is said to be Pareto optimal if $\m{x}$ is not dominated by any points in $\mathbb{R}^n$. Similarly, $\m{x}$ is locally Pareto optimal if $\m{x}$ is not dominated by any points in a neighborhood of $\m{x}$. The Pareto set of this problem consists of all Pareto optimal points, and the Pareto front is the image of the Pareto set. In the context of deep MTL, $\m{x}$ represents the parameters of a neural network instance and each $f_i(\m{x})$ represents one learning objective, e.g., a certain classification loss.

Similar to single-objective optimization, solving for local Pareto optimality is better established than global Pareto optimality. A standard way is to run gradient-based methods to solve for local Pareto optimality then prune the results. Hillermeier et al. \yrcite{hillermeier01} describes the following necessary condition:
\begin{definition}[\citealt{hillermeier01}]\label{def:stationary}
Assuming each $f_i(\m{x})$ is continuously differentiable, a point $\m{x}$ is called Pareto stationary if there exists $\m{\alpha}\in\mathbb{R}^m$ such that $\alpha_i\geq0$, $\sum_{i=1}^{m}\alpha_i=1$, and $\sum_{i=1}^{m}\alpha_i\nabla f_i(\m{x})=\m{0}$.
\end{definition}

\begin{proposition}[\citealt{hillermeier01}]\label{thm:kkt} All Pareto optimal points are Pareto stationary.
\end{proposition}

Once a Pareto optimal solution $\m{x}^*$ is found, previous papers \cite{hillermeier01b, martin18,schulz18} have proven a strong result revealing the first-order approximation of the local, continuous Pareto set:
\begin{proposition}[\citealt{hillermeier01b}]\label{thm:exploration}
Assuming that $\m{f}(\m{x})$ is smooth and $\m{x}^*$ is Pareto optimal, consider any smooth curve $\m{x}(t):(-\epsilon,\epsilon)\rightarrow\mathbb{R}^n$ in the Pareto set and passing $\m{x}^*$ at $t=0$, i.e., $\m{x}(0)=\m{x}^*$, then $\exists\m{\beta}\in\mathbb{R}^m$ such that:
\begin{equation}\label{eq:core}
    \m{H}(\m{x}^*)\m{x}'(0)=\nabla \m{f}(\m{x}^*)^\top\m{\beta}
\end{equation}
where $\m{H}(\m{x}^*)$ is defined as
\begin{equation}
    \m{H}(\m{x}^*)=\sum\nolimits_{i=1}^m\alpha_i\nabla^2 f_i(\m{x}^*)
\end{equation}
and $\alpha_i$ is given by Definition~\ref{def:stationary}.
\end{proposition}
In other words, in the Pareto set, for any smooth curve passing $\m{x}^*$, $\m{H}(\m{x}^*)$ transforms its tangent at $\m{x}^*$ to a vector in the space spanned by $\{\nabla f_i(\m{x}^*)\}$. By gradually changing the curve, its tangent sweeps the tangent plane of the Pareto set at $\m{x}^*$. Essentially, the theorem states that $\m{H}(\m{x}^*)$ connects the whole tangent plane to the column space of $\nabla \m{f}(\m{x}^*)^\top$. Note that, however, this theorem is not directly applicable to MTL because of its requirement of full Hessians.
\section{Efficient Pareto Set Exploration}

Given an initial $\m{x}_0\in\mathbb{R}^n$, our algorithm is executed in two phases: phase 1 uses gradient-based methods to generate a Pareto stationary solution $\m{x}_0^*$ from $\m{x}_0$. It then computes a few exploration directions to spawn new $\{\m{x}_i\}$. We execute phase 1 recursively by feeding it with a newly generated $\m{x}_i$. Phase 2 constructs continuous Pareto sets: we first build a local linear subspace at each Pareto stationary solution by linearly combining its exploration directions. We then check whether two local Pareto fronts collide and stitch them to form a larger continuous set. The major challenge brought by deep MTL is that $\mathbb{R}^n$ is the space of neural network parameters. Therefore, it is computationally prohibitive to explicitly calculate Hessian matrices. We describe phase 1 below and phase 2 will be explained in Section \ref{sec:continuous}.

\subsection{Gradient-Based Optimization}\label{sec:optimizer}
Our algorithm is compatible with any gradient-based local optimization methods as long as they can return a Pareto stationary solution from any initial $\m{x}\in\mathbb{R}^n$. A standard method in MTL is to minimize a weighted sum of objectives with stochastic gradient descent (SGD) \cite{kokkinos17,chen17,kendall18}. Recent papers \cite{sener18,lin19} also proposed to determine a gradient direction online by solving a small convex problem. Essentially, they minimize a loss by combining gradients with fixed or adaptive weights.

\subsection{First-Order Expansion}\label{sec:expand}
Once a Pareto stationary point $\m{x}^*_0$ is found, we explore its local Pareto set by spawning new points $\{\m{x}_i\}$. This is decomposed into two steps: computing $\m{\alpha}$ in Definition~\ref{def:stationary} at $\m{x}^*_0$ and estimating  $\{\m{v}_i\}$, the basis directions of the tangent plane, from Proposition \ref{thm:exploration}. The new points $\{\m{x}_i\}$ are then computed by $\m{x}_i=\m{x}^*_0+s\m{v}_i$ where $s$ is an empirical step size whose choice will be discussed in our experiments.

We acquire $\m{\alpha}$ at $\m{x}^*_0$ by solving the following convex problem \cite{desideri12}, as suggested by Sener and Koltun \yrcite{sener18}:
\begin{equation}\label{eq:cvx}
\begin{aligned}
    \min_{\m{\alpha}}\quad&\|\sum\nolimits_{i=1}^m \alpha_i\nabla f_i(\m{x}^*_0)\|_2\\
    s.t.\quad&\m{\alpha}\geq\m{0},\quad\sum\nolimits_{i=1}^m\alpha_i=1
\end{aligned}
\end{equation}
Note that the objective can be written as a quadratic form of dimension $m$. Since $m$ is typically very small, solving it takes little time even for large neural networks.

Given $\m{\alpha}$, finding $\{\m{v}_i\}$ on the tangent plane at $\m{x}^*_0$ can be transformed to finding a solution $(\m{v},\m{\beta})$ from Equation (\ref{eq:core}):
\begin{equation}\label{eq:nullspace}
    \m{H}(\m{x}^*_0)\m{v}=\nabla\m{f}(\m{x}^*_0)^\top\m{\beta}
\end{equation}
When $n$ is small, we can apply classic $O(n^3)$ methods like Gram-Schmidt process or QR decomposition. However, directly applying them in deep MTL is difficult for two reasons: first, $\m{x}^*_0$ is rarely a true Pareto stationary solution because of the early termination in training to avoid overfitting. Second, and more importantly, the large parameter space makes any $O(n^3)$ method prohibitive.

To address the first issue, we propose a variant to Problem (\ref{eq:cvx}) to find $\m{\alpha}$ as well as a correction vector $\m{c}$:
\begin{equation}\label{eq:correction}
\begin{aligned}
\min_{\m{\alpha},\m{c}}\quad&\|\m{c}\|_2\\
s.t.\quad&\m{\alpha}\geq\m{0},\quad\sum\nolimits_{i=1}^m\alpha_i=1\\
&\sum\nolimits_{i=1}^{m}\alpha_i(\nabla f_i(\m{x}^*_0)-\m{c})=\m{0}
\end{aligned}
\end{equation}
In other words, we seek the minimal modification to the gradients such that if we use $\nabla f_i(\m{x}^*_0)-\m{c}$ as if they were the true gradients, $\m{x}^*_0$ would be Pareto stationary. It is easy to show that solving this new optimization problem brings little overhead to the original problem (see supplemental material for the proof):
\begin{proposition}\label{thm:correction}
Let $\m{\alpha}^*$ be the solution to Problem (\ref{eq:cvx}), then the solution to Problem \ref{eq:correction} is $(\m{\alpha},\m{c})=(\m{\alpha}^*,\nabla\m{f}(\m{x}^*_0)^\top\m{\alpha}^*)$.
\end{proposition}

To address scalability, we consider the following sparse linear system with unknowns $\m{v}$:
\begin{equation}\label{eq:sample}
    \m{H}(\m{x}^*_0)\m{v}=(\nabla\m{f}(\m{x}^*_0)^\top-\m{c}\m{1}^\top)\m{\beta}
\end{equation}
where $\m{1}$ is an $m$-dimensional column vector with all elements equal to $1$ and $\m{\beta}\in\mathbb{R}^m$ is randomly sampled. In other words, we solve a linear system with the right-hand side sampled from the space spanned by $\{\nabla f_i(\m{x}^*_0)-\m{c}\}$. Solving such a large linear system in MTL requires an efficient matrix solver. We propose to use Krylov subspace iteration methods because they are matrix-free and iterative solvers, allowing us to solve the system without complete Hessians and terminate with intermediate results. In our experiment, we choose to use the minimal residual method (MINRES), a classic Krylov subspace method designed for symmetric indefinite matrices \cite{choi11}.

We now discuss MINRES in more detail to better explain why it is the right tool for this problem. The time complexity of MINRES depends on the time spent on each iteration and the number of iterations. The cost of each iteration is dominated by calculating $\m{H}\m{v}$ for arbitrary $\m{v}$, which is in general $O(n^2)$. However, it is well known that Hessian-vector products can be implemented in $O(n)$ time on computational graphs \cite{pearlmutter94}, giving us the first strong reason to use MINRES. Analyzing the number of iterations is hard because it heavily depends on the rarely available eigenvalue distribution. In practice, MINRES is known to converge very fast for systems with fast decay of eigenvalues \cite{fong12}. In our experiments, we specify a maximum number of iterations $k$. We observed that $k=50$ was usually sufficient to generate good exploration directions even for networks with millions of parameters. Note that early termination in MINRES still returns meaningful results because the residual error is guaranteed to decrease monotonically with iterations.

To summarize, the efficiency of our exploration algorithm comes from two sources: exploration on the tangent plane and early termination from a matrix-free, iterative solver. The time cost of getting one tangent direction is $O(kn)$, which scales linearly to the network size.

\subsection{The Full Algorithm}
We now state the complete algorithm for Pareto set exploration in Algorithm \ref{alg:explore}. It takes as input a seed network and spawns $N$ Pareto stationary networks in a breadth-first style. Any networks put in the queue are returned by \texttt{ParetoOptimize} (Section \ref{sec:optimizer}) and therefore Pareto stationary by design. When such a network is popped out from the queue, \texttt{ParetoExpand} generates $K$ exploration directions (Section \ref{sec:expand}) and spawns $K$ child networks. The algorithm then calls \texttt{ParetoOptimize} to refine these networks before appending them to the queue, and terminates after $M$ Pareto stationary networks are collected.
\begin{algorithm}[tb]
    \caption{Efficient Pareto Set Exploration}
    \label{alg:explore}
\begin{algorithmic}
    \STATE {\bfseries Input:} a random initial neural network $\m{x}_0\in\mathbb{R}^n$
    \STATE {\bfseries Output:} $N$ Pareto stationary networks
    \STATE $\m{x}^*_0\leftarrow$\texttt{ParetoOptimize}($\m{x}_0$)
    \STATE Initialize a queue $q\leftarrow[\m{x}^*_0]$
    \STATE Initialize an empty list to store the output: $output\leftarrow\emptyset$
    \REPEAT
    \STATE Pop a neural network $\m{x}^*$ from $q$
    \FOR{$i=1$ {\bfseries to} $K$}
        \STATE $\m{v}_i\leftarrow$\texttt{ParetoExpand}($\m{x}^*$)
        \STATE $\m{v}_i/{=}\|\m{v}_i\|_2$
        \STATE $\m{x}_i\leftarrow\m{x}^*+s\m{v}_i$
        \STATE $\m{x}^*_i\leftarrow$\texttt{ParetoOptimize}($\m{x}_i$)
        \IF{No points in $output$ dominates $\m{x}^*_i$}
            \STATE Append $\m{x}^*_i$ to $q$
            \STATE Append $(\m{x}^*_i,\m{f}(\m{x}^*_i),\nabla\m{f}(\m{x}^*_i),\m{x}^*)$ to $output$
        \ENDIF
    \ENDFOR
\UNTIL{The size of $output$ reaches $N$}
\end{algorithmic}
\end{algorithm}

For each output network, we also return the objectives, the gradients, and a reference to its parent. This information is mostly used to construct a continuous linear subspace approximating the local Pareto set, which we will describe in Section \ref{sec:continuous}. Another usage is to remove the sign ambiguity in $\m{v}_i$: by definition, both $\m{v}_i$ and $-\m{v}_i$ are on the tangent plane, and an arbitrary choice can lead to a retraction instead of the desired expansion in the Pareto set. In this case, one can use $\m{f}(\m{x}_i)-\m{f}(\m{x}^*)=\m{f}(\m{x}^*+s\m{v}_i)-\m{f}(\m{x}^*)\approx s\nabla\m{f}(\m{x}^*)\m{v}_i$ to predict the changes in the objectives and rule out the undesired direction.

When Algorithm \ref{alg:explore} is applied to MTL, it is worth noting that \texttt{ParetoOptimize} and \texttt{ParetoExpand} rarely return the precise solutions because of stochasticity, early termination, and local minima. As a result, good choices of hyperparameters plays an important role. We discussed in more detail two crucial hyperparameters ($k$ and $s$) and reported the ablation study in Section \ref{sec:result}.
\section{Continuous Parametrization}\label{sec:continuous}
In this section, we describe a post-processing step that builds a continuous approximation to the local Pareto set based on the discrete points $\{\m{x}^*_i\}$ returned by Algorithm \ref{alg:explore}. For each $\m{x}^*_i$, we collect its $K$ children $\{\m{x}^*_{i_1},\cdots,\m{x}^*_{i_K}\}$ and assign a continuous variable $r_{i\rightarrow i_j}\in[0,1]$ to a vector $\m{v}_{i\rightarrow i_j}=\m{x}^*_{i_j}-\m{x}^*_i,j=1,2,\cdots,K$. The local Pareto set at $\m{x}^*_i$ is then constructed by
\begin{equation}
    S(\m{x}^*_i)=\{\m{x}^*_i+\sum_{i=1}^K r_{i\rightarrow i_j}\m{v}_{i\rightarrow i_j}| r_{i\rightarrow i_j}\geq0,\sum_{i=1}^K r_{i\rightarrow i_j}\leq1\}
\end{equation}
In other words, $S(\m{x}^*)$ is the convex hull of $\m{x}^*_i$ and its children $\{\m{x}^*_{i_1},\cdots,\m{x}^*_{i_K}\}$. This construction is justified by the fact that a linear combination of tangent vectors is still on the tangent plane. As a special case, when there are only 2 objectives and $K=1$, $\{\m{x}^*\}$ forms a chain, and therefore $S=\cup_i S(\m{x}^*_i)$ becomes a piecewise linear set in $\mathbb{R}^n$.

It is possible that two continuous families can collide in the objective space, creating a larger continuous Pareto front. In this case, we create a stitching point in both families and crop solutions dominated by the other family. By repeatedly applying this idea, a single continuous Pareto front covering all families can possibly be created, providing the ultimate solution to continuous traversal in the whole Pareto front. We illustrate this idea on MultiMNIST with our experimental results in Section \ref{sec:exp:continuous}.

Since the continuous approximation interpolates different tangent directions, having more directions can enrich the coverage of the continuous set and offer more options to users. It is therefore natural to ask whether the set of tangent directions discovered in the last section could be augmented even further by adding more directions without downgrading the quality of the Pareto front. For the special case of two objectives ($m=2$), it turns out that we can augment the set of known tangent directions with a \textit{null vector} of the Hessian matrix, as stated in the following proposition:
\begin{proposition}\label{thm:null}
Assuming $\m{f}(\m{x}):\mathbb{R}^n\rightarrow\mathbb{R}^2$ is sufficiently smooth. Let $\m{x}^*$ be a Pareto optimal point and consider a curve $\m{c}_{\m{d}}(t):\mathbb{R}\rightarrow\mathbb{R}^2$ defined as $\m{c}_{\m{d}}(t)=\m{f}(\m{x}^*+t\m{d})$. If $\m{x}(t):(-\epsilon,\epsilon)\rightarrow\mathbb{R}^2$ is any smooth curve in Proposition \ref{thm:exploration} that satisfies $\m{H}(\m{x}^*)\m{x}'(0)\neq\m{0}$, then for any $\m{u}\in\mathbb{R}^n$:

1) $\m{c}_{\m{x}'(0)}$ and $\m{c}_{\m{x}'(0)+\m{u}}$ have the same value and tangent direction $(-\alpha_2,\alpha_1)$ at $t=0$;

2) Furthermore, if $\m{u}$ is a null vector of $\m{H}(\m{x}^*)$, i.e., $\m{H}(\m{x}^*)\m{u}=\m{0}$, then $\m{u}$ is not parallel to $\m{x}'(0)$, and  $\m{c}_{\m{x}'(0)}(t)$ and $\m{c}_{\m{x}'(0)+\m{u}}(t)$ have the same curvature at $t=0$.
\end{proposition}
In this proposition, $\m{c}_{\m{d}}(t)$ is a parametrized 2D curve: it considers a straight-line trajectory in $\mathbb{R}^n$ that passes $\m{x}^*$ in the direction of $\m{d}$ and uses $\m{f}$ to map this trajectory to the space of $\mathbb{R}^2$, generating a 2D curve. This proposition states that if a tangent direction $\m{v}$ is known and if we also have a null vector $\m{u}$, then the two curves $\m{c}_{\m{v}}$ and $\m{c}_{\m{v}+\m{u}}$ are very similar at $\m{x}^*$ in the sense that they share the same value, gradients, and curvature. This means that for each tangent direction $\m{v}$ found in the previous section, $\m{v}+\m{u}$ can also be used as a backbone direction together with $\m{v}$ for continuous parametrization without downgrading the quality of the reconstructed Pareto front.

While this proposition is generally not applicable to real problems due to its need for null vectors, it still has interesting theoretical implications: the fact that $\m{c}_{\m{v}+\m{u}}$ and $\m{c}_{\m{v}}$ share the same gradients should not be surprising as $\m{v}+\m{u}$ also satisfies Equation~(\ref{eq:nullspace}), but it is less obvious to see that they actually share the same curvature at $\m{f}(\m{x}^*)$, which we illustrate in Section \ref{sec:exp:continuous} and will prove in our supplemental material. In practice, we observed that neural networks typically have a Hessian matrix with a null space whose dimension is much higher than $m$. This means a very large set of bases, while not often accessible in real problems, can in theory be used to greatly enrich the Pareto set.
\section{Experimental Results}\label{sec:result}

\subsection{Datasets, Metrics, and Baselines}\label{sec:exp:setup}
We applied our method to five datasets in three categories: 1) MultiMNIST \cite{sabour17} and its two variants FashionMNIST \cite{xiao17} and MultiFashionMNIST, which are medium-sized datasets with two classification tasks; 2) UCI Census-Income \cite{kohavi96}, a medium-sized demographic dataset with three binary prediction tasks; 3) UTKFace \cite{zhang17}, a large dataset of face images. We used LeNet5 \cite{lecun98} (22,350 parameters) for MultiMNIST and its variants, two-layered multilayer perceptron (158,598 parameters) for UCI Census-Income, and ResNet18 \cite{he16} (tens of millions of parameters) for UTKFace. Please refer to our supplemental material for more information about the network architectures, task descriptions, and implementation details in each dataset.

\begin{figure}[!b]
\vskip 0.2in
\begin{center}
\centerline{\includegraphics[width=\columnwidth]{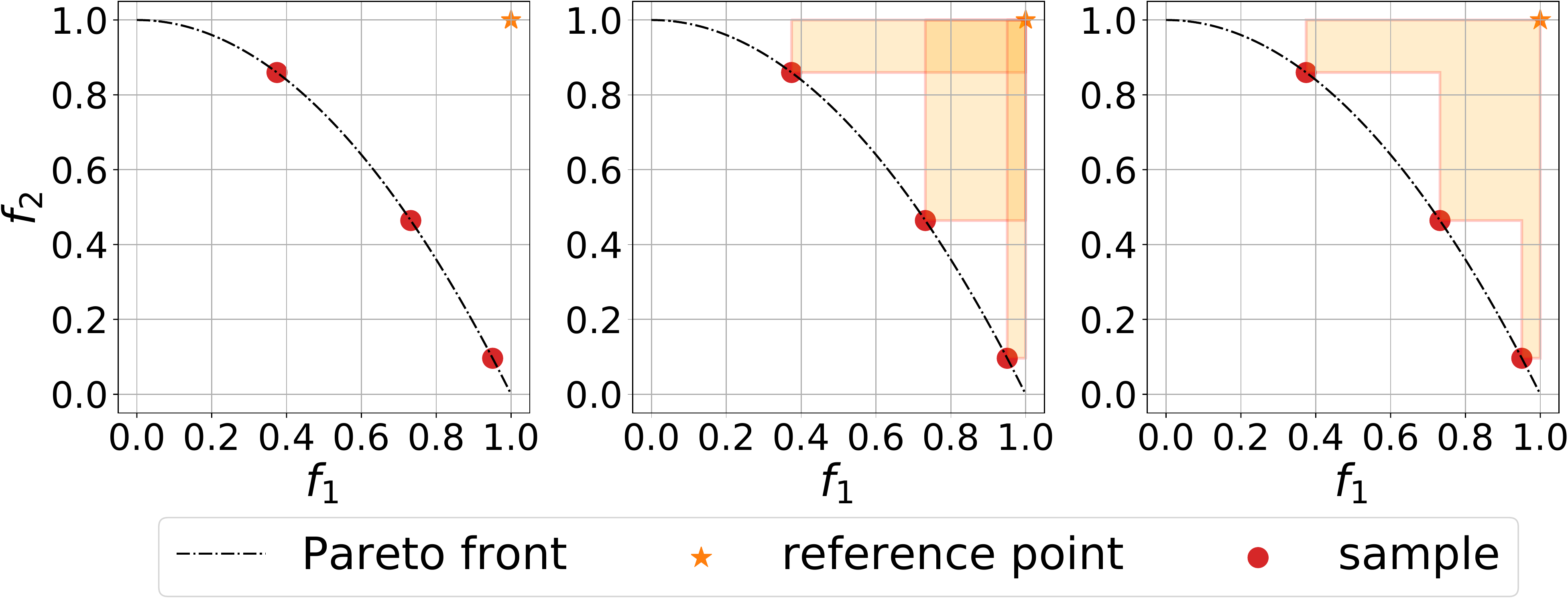}}
\caption{Definition of hypervolume. Given a set of sample points (red circles) in $\mathbb{R}^m$, the hypervolume is computed by picking a reference point (orange star), creating axis-aligned rectangles from each point, and calculating the size of their union (orange polygon).}
\label{fig:hv}
\end{center}
\vskip -0.2in
\end{figure}

We measure the performance of a method by two metrics: the time cost and the hypervolume \cite{zitzler99}. We measure the time cost by counting the evaluations of objectives, gradients, and Hessian-vector products. The hypervolume metric, explained in Figure \ref{fig:hv}, is a classic MOO metric for measuring the quality of exploration. More concretely, this metric takes as input a set of explored solutions in the objective space and returns a score. Larger hypervolume score indicates a better Pareto front. Using the two metrics, we define that a method is more efficient if, within the same time budget, it generates a Pareto front with a larger hypervolume, or equivalently, if it generates the Pareto front with a similar hypervolume but within shorter time. For all figures in this section, we use the same random seed whenever possible and report results from more random seeds in the supplemental material.

\begin{figure}[!t]
\vskip 0.2in
\begin{center}
\centerline{\includegraphics[width=\columnwidth]{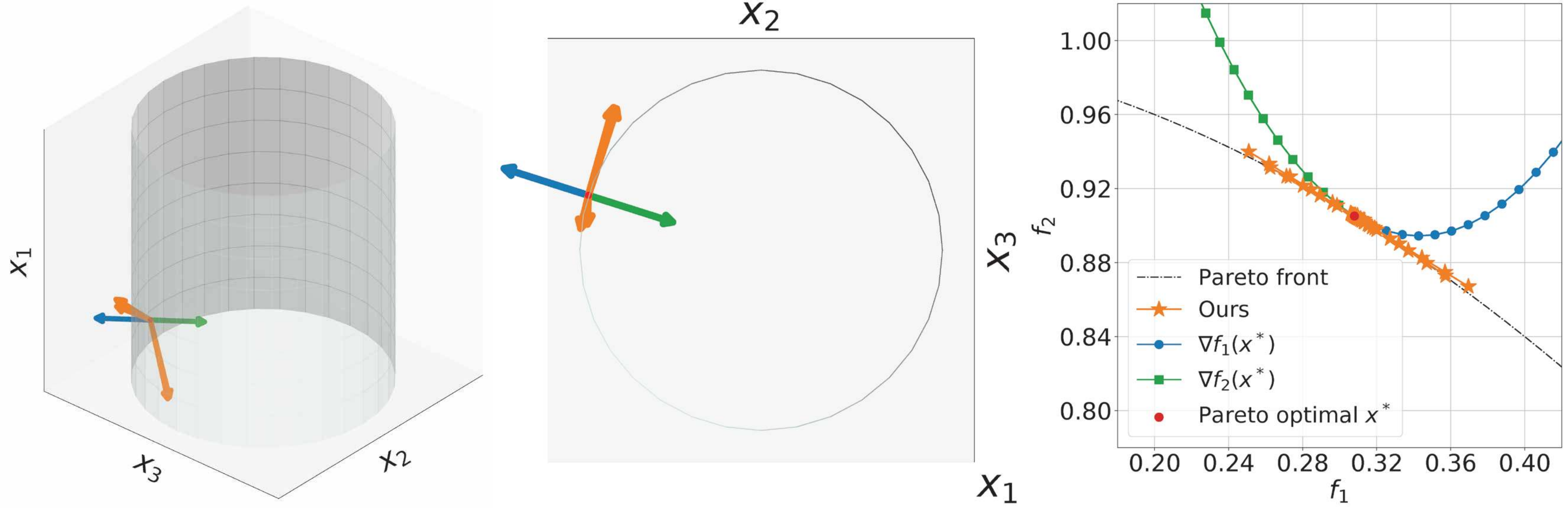}}
\caption{Comparisons of different exploration directions at a Pareto optimal solution $\m{x}^*$ (red circle). Left: the analytic Pareto set (the cylindrical surface) of ZDT2-variant, the gradients $\nabla f_1(\m{x}^*)$ (blue) and $\nabla f_2(\m{x}^*)$ (green), and our exploration directions $\{\m{v}_i\}$ (orange) predicted by MINRES. Middle: a top-down view to show ours are almost tangent to the Pareto set. Right: plots of $\m{f}(\m{x}^*+s\m{d})$ where $s\in[-0.1,0.1]$ and $\m{d}$ is $\nabla f_1(\m{x}^*)$ (blue circles), $\nabla f_2(\m{x}^*)$ (green squares), and our directions (orange stars).}
\label{fig:zdt_tangent}
\end{center}
\vskip -0.2in
\end{figure}

Our method is not directly comparable to any baselines because no prior work aims to recover a continuous Pareto front in MTL. Instead, we devised two experiments, which we call the sufficiency and necessity tests, to show its effectiveness (Section \ref{sec:exp:expand}). In the sufficiency test, we consider four previous methods: GradNorm \cite{chen17}, Uncertainty \cite{kendall18}, MGDA \cite{sener18}, and ParetoMTL \cite{lin19}. These methods aim at pushing an initial guess to one or a few discrete Pareto optimal solutions. For them, we show that our Pareto expansion procedure is a fast yet powerful complement by comparing the time and hypervolume before and after running it as a post-processing step. We call this experiment the sufficiency test as it demonstrates our method is able to quickly explore Pareto sets and Pareto fronts.

Our necessity test, which focuses on the value of the tangent directions in exploring Pareto fronts, deserves some discussions on its baselines. There is a trivial baseline for Pareto expansion: rerunning an SGD-based method from scratch to optimize a perturbed weight combination of objectives. Since each new run requires full training, our method clearly dominates this baseline ($30$ times faster on MultiMNIST). Another trivial baseline is to use a random direction instead of the tangent direction for Pareto expansion. We tested this idea but do not include it in our experiments as its performance is significantly worse than any other methods, which is understandable due to the high dimensionality of neural network parameters: with the increase of dimensionality, the chance of a random guess still staying on the tangent plane decays exponentially. The baseline we considered in this experiment is WeightedSum, which runs SGD from the last Pareto optimal solution but with weights on objectives different from the weights used in training. Specifically, we choose weights from one-hot vectors for each task as well as a vector assigning equal weights to every task. We call this experiment the necessity test as we use this experiment to establish that the choice of expansion strategies is not arbitrary, and tangent directions are indeed the source of efficiency in our method.

\begin{figure}[!t]
\vskip 0.2in
\begin{center}
\centerline{\includegraphics[width=\columnwidth]{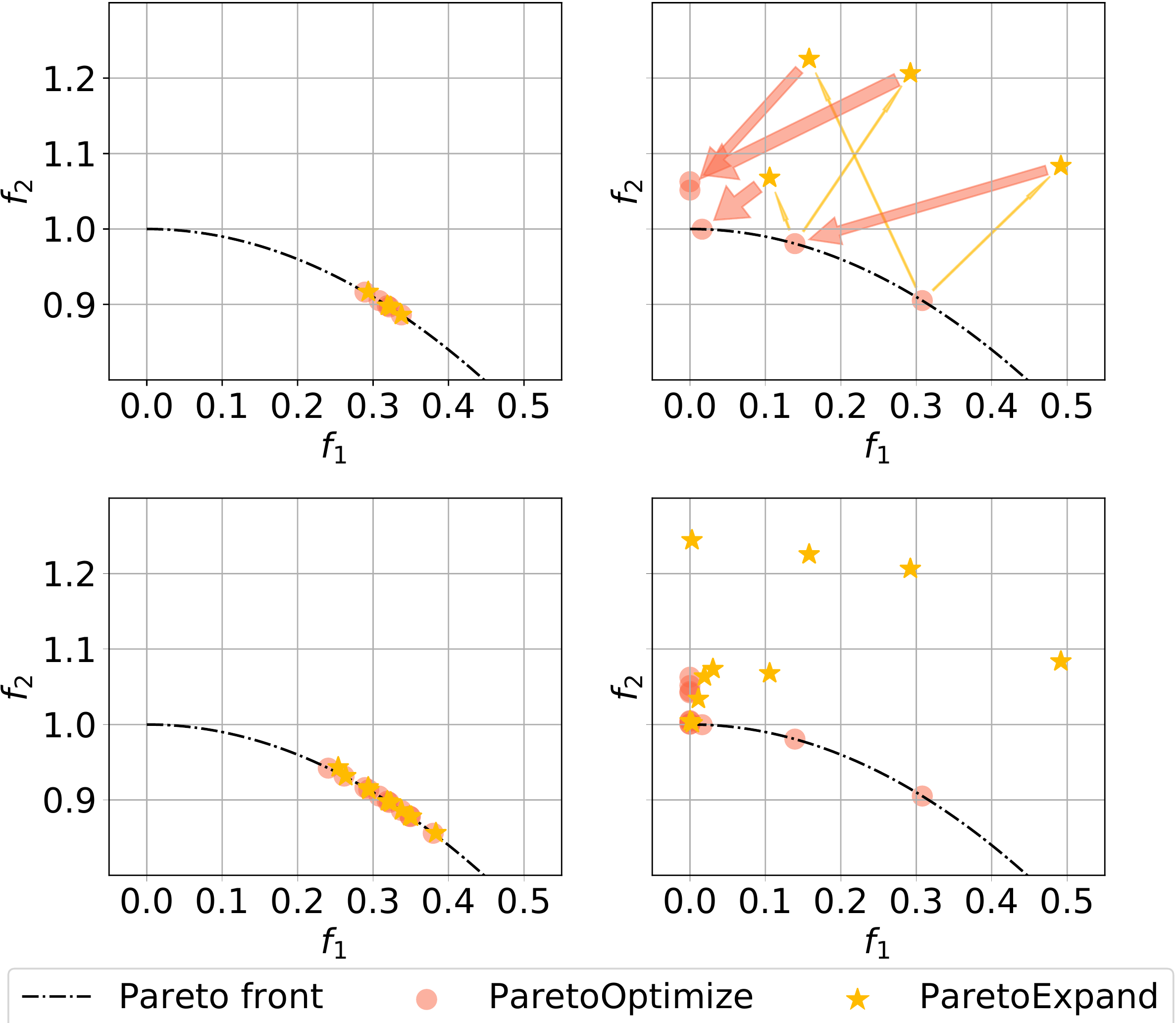}}
\caption{Comparisons of two expansion strategies in Algorithm \ref{alg:explore}. Starting with a given Pareto optimal point (red circle), the algorithm iteratively calls \texttt{ParetoExpand} (orange arrows from circles to stars) and \texttt{ParetoOptimize} (red arrows from stars to circles), generating a series of explored points (orange stars) and Pareto optimal solutions (red circles). Arrow thickness indicates the time cost of each function. Top row: expansion using our predicted tangent directions (top left) versus using gradients (top right). Bottom row: running both strategies until 10 Pareto optimal points were collected.}
\label{fig:zdt_expand}
\end{center}
\vskip -0.2in
\end{figure}

\subsection{Synthetic Examples}\label{sec:exp:synthetic}

\subsubsection{ZDT2-variant}
Our first example, ZDT2-variant, was originated from ZDT2 \cite{zitzler00}, a classic benchmark problem in multi-objective optimization with $n=3$ and $m=2$. Both the Pareto set and the Pareto front of this example can be computed analytically. This makes ZDT2-variant an ideal example for visualizing Proposition \ref{thm:exploration} and Algorithm \ref{alg:explore}. Figure \ref{fig:zdt_tangent} compares the gradients to our tangent directions when used to explore the Pareto front. We used MINRES with $k=1$ to solve 5 tangent directions. It can be seen that our directions are much closer to the Pareto set and tracked the true Pareto front much better than the gradients. We further compare their performances in Algorithm \ref{alg:explore} with MGDA \cite{desideri12,sener18} as the optimizer in Figure \ref{fig:zdt_expand}. This figure shows that the gradients expanded the neighborhood not on the Pareto set but to the dominated interior, resulting in a much more expensive correction step to follow. On the other hand, expanding with our predicted tangents steadily grew the solution set along the Pareto front.

\subsubsection{MultiMNIST Subset}
\begin{figure}[htb]
\vskip 0.2in
\begin{center}
\centerline{\includegraphics[width=\columnwidth]{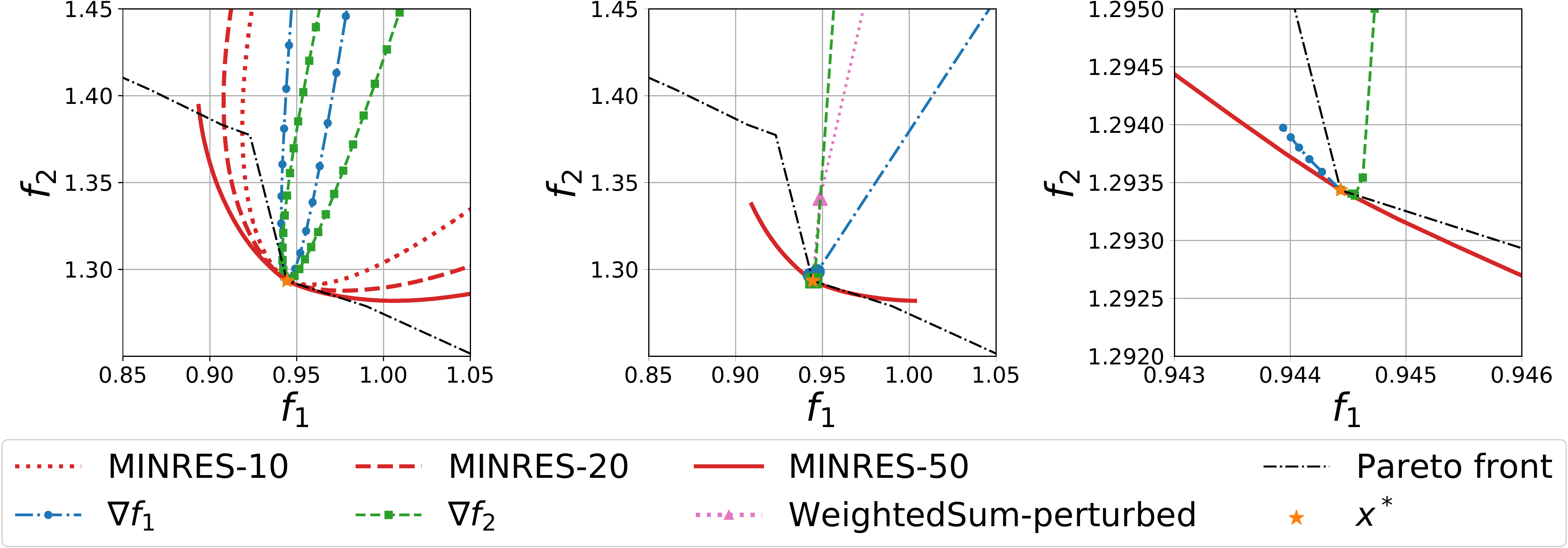}}
\caption{Comparisons of expansion strategies on MultiMNIST subset. Left: trajectories of different expansion strategies in the objective space. Curves closer to the Pareto front mean better expansion. Middle: trajectories generated by running SGD to minimize $f_1$ (blue circles), $f_2$ (green squares), or a weighted combination (pink triangles) within the same time budget as MINRES-50 (red). Right: a zoom-in version showing that tiny step sizes have to be used by SGD to avoid deviating from the Pareto front too much.}
\label{fig:m3_expand}
\end{center}
\vskip -0.2in
\end{figure}

To understand the behavior of our algorithm when neural networks are involved, we picked a subset of $2048$ images from MultiMNIST and trained a simplified LeNet \cite{lecun98} with 1500 parameters to minimize two classification errors. We generated an empirical Pareto front by optimizing the weighted sum of the two objectives with varying weights. We then picked a Pareto optimal $\m{x}^*$ and visualized trajectories generated by traversing along gradients and the approximated tangents after $10$, $20$, and $50$ iterations of MINRES (Figure \ref{fig:m3_expand} left). Just as in ZDT2-variant, our approximated tangents tracked the Pareto front much more closely. We then compared using approximated tangents after 50 iterations of MINRES (MINRES-50) to the WeightedSum baseline (Section \ref{sec:exp:setup}) after 50 iterations of SGD. The two methods had roughly the same time budgets, and MINRES-50 outperformed the WeightedSum baseline in that it explored a much wider Pareto front (Figure \ref{fig:m3_expand} middle). Specifically, its advantage comes from a much larger step size enabled by the approximated tangents (Figure \ref{fig:m3_expand} right).

\subsection{Pareto Expansion}\label{sec:exp:expand}
\begin{figure}[tb]
\vskip 0.2in
\begin{center}
\centerline{\includegraphics[width=\columnwidth]{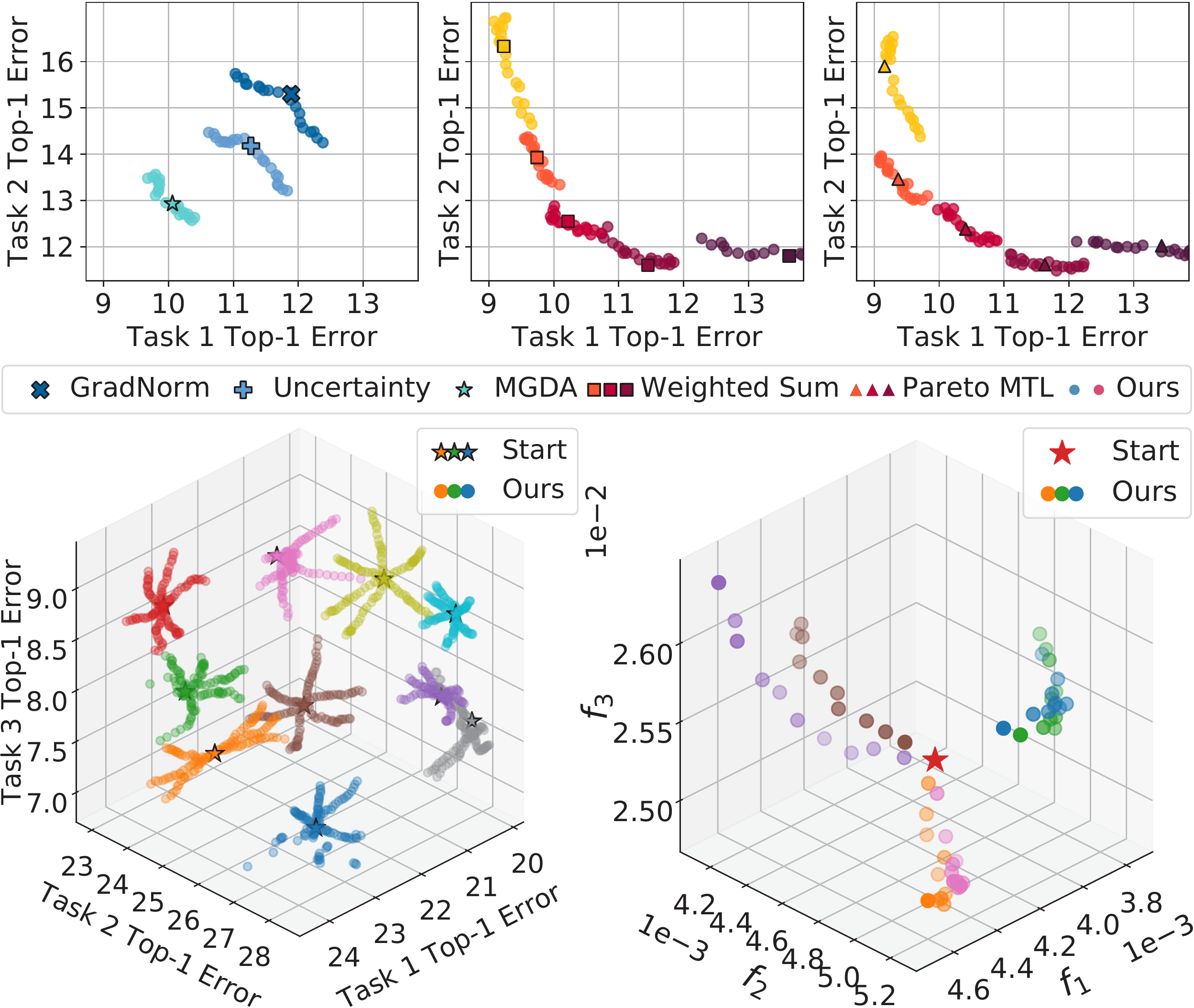}}
\caption{Using Pareto expansion to grow dense Pareto fronts (colorful circles) from discrete solutions generated by baselines on MultiMNIST and its variants (top row), UCI Census-Income (bottom left), and UTKFace (bottom right). Points expanded from the same discrete solution have the same color.}
\label{fig:complement}
\end{center}
\vskip -0.2in
\end{figure}

We first conducted the sufficiency test described in Section \ref{sec:exp:setup} to analyze Pareto expansion, the core of our algorithm. We ran ParetoMTL, the state of the art, on all datasets to generate discrete seeds for Pareto expansion. Moreover, for smaller datasets (MultiMNIST and its variants), we also ran the other baselines for a more thorough analysis. Compared to the time cost of generating discrete solutions (Table \ref{tb:complement} column 2), our Pareto expansion only used a small fraction of the training time (Table \ref{tb:complement} column 4) but generated much denser Pareto fronts (Figure \ref{fig:complement} and Table \ref{tb:complement} column 5). This experiment, as a natural extension to the synthetic experiments, confirms the efficacy of Pareto expansion on large neural networks and datasets.

\begin{figure}[!b]
\vskip 0.2in
\begin{center}
\centerline{\includegraphics[width=\columnwidth]{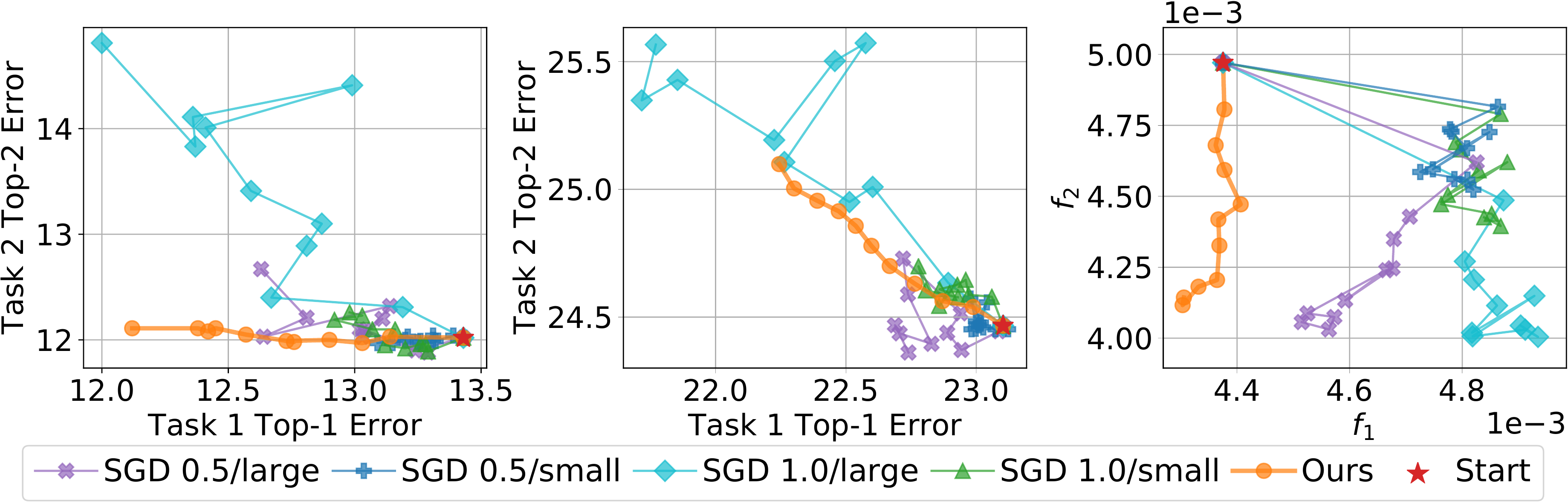}}
\caption{Comparisons of two expansion methods (ours and running SGD with a weighted sum) from a given Pareto optimal network. We display results on MultiMNIST (left), UCI Census-Income (middle), and UTKFace (right). In all figures, lower left regions mean better solutions. All SGD methods are labeled with preference on task 1/the type of learning rates (large or small). UCI Census-Income and UTKFace have three objectives and we show results from considering $f_1$ and $f_2$ only. Results on FashionMNIST and MultiFashionMNIST and other combinations of objectives in UCI Census-Income and UTKFace can be found in our supplemental material.}
\label{fig:fine_tune}
\end{center}
\vskip -0.2in
\end{figure}

The sufficiency test has established that our expansion method has a positive effect on discovering more solutions. However, one can still argue there could be simpler expansion strategies that are as good as ours. It remains to show that the benefit indeed comes from approximated tangent directions. We verified this with the necessity test described in Section \ref{sec:exp:setup}, which directly compared our Pareto expansion to the WeightedSum expansion strategy. Starting with the same seed solution, we gave both methods the same time budget, so the area of their expansions directly reflected their performances. We display the results on MultiMNIST, UCI Census-Income, and UTKFace in Figure \ref{fig:fine_tune}. New solutions were generated after each run of MINRES in our method and after each epoch in WeightedSum. We provide more results in the supplementary material. We see from these experiments that our method discovered solutions that clearly dominated what WeightedSum returned on 4 out of the 5 datasets except UCI Census-Income. From this experiment, we conclude the tangent directions in Pareto expansion are indeed the core reason for the good performance of our algorithm.

The effectiveness of our Pareto expansion method can also be understood by noticing it uses higher-order derivatives than previous work for determining the optimal expansion directions. Consider the three possible methods for the task of expanding the local Pareto set from a known Pareto optimal solution $\m{x^*}$: simply retraining the neural network from scratch with a different initial guess reuses nothing from $\m{x^*}$; rerunning SGD from $\m{x^*}$ leverages the first-order gradient information at $\m{x^*}$; our method exploits both the first-order and the second-order information at $\m{x^*}$ and therefore is the most effective among the three.

It is worth mentioning that our Pareto expansion strategy is still a local optimization method, meaning that it inevitably suffers from being trapped in local minima. As a result, there is no theoretical guarantee on the resulting Pareto fronts being globally Pareto optimal. We alleviate this issue by exploring from multiple Pareto optimal solutions returned by previous methods and stitching them together, which we will explain shortly in the next section.

\begin{table}[!t]
\caption{A summary of the improvement brought by calling Pareto expansion from solutions generated by baselines. TRAIN: the training time used by each baseline, measured by the aggregated number of evaluations of objectives, gradients, and Hessian-vector products; HV: the hypervolume of the solution at the end of training; EXPAND: the time cost of our Pareto expansion; NEW HV: the hypervolume after expansion. Larger hypervolume is better.}

\label{tb:complement}
\vskip 0.15in
\begin{center}
\begin{small}
\begin{sc}
\begin{tabular}{rcccc}
\toprule
MultiMNIST & train & hv & expand & new hv \\
\midrule
GradNorm & 21150 & 7.463 & 4520 & 7.628 \\
Uncertainty & 21150 & 7.615 & 4520 & 7.756 \\
MGDA & 21150 & 7.831 & 4520 & 7.896 \\
WeightedSum & 70500 & 8.019 & 22600 & 8.034 \\
ParetoMTL & 106281 & 8.025 & 22600 & 8.046 \\
\midrule
UCI & train & hv & expand & new hv \\
\midrule
WeightedSum & 467400 & 5.685 & 165600 & 5.725 \\
ParetoMTL & 934888 & 5.642 & 165600 & 5.675 \\
\midrule
Face & train & hv & expand & new hv \\
\midrule
ParetoMTL & 35568 & 2.257 & 9920 & 5.030 \\
\bottomrule
\end{tabular}
\end{sc}
\end{small}
\end{center}
\vskip -0.1in
\end{table}

\subsection{Continuous Parametrization}\label{sec:exp:continuous}

\begin{figure}[!tb]
\vskip 0.2in
\begin{center}
\centerline{\includegraphics[width=\columnwidth]{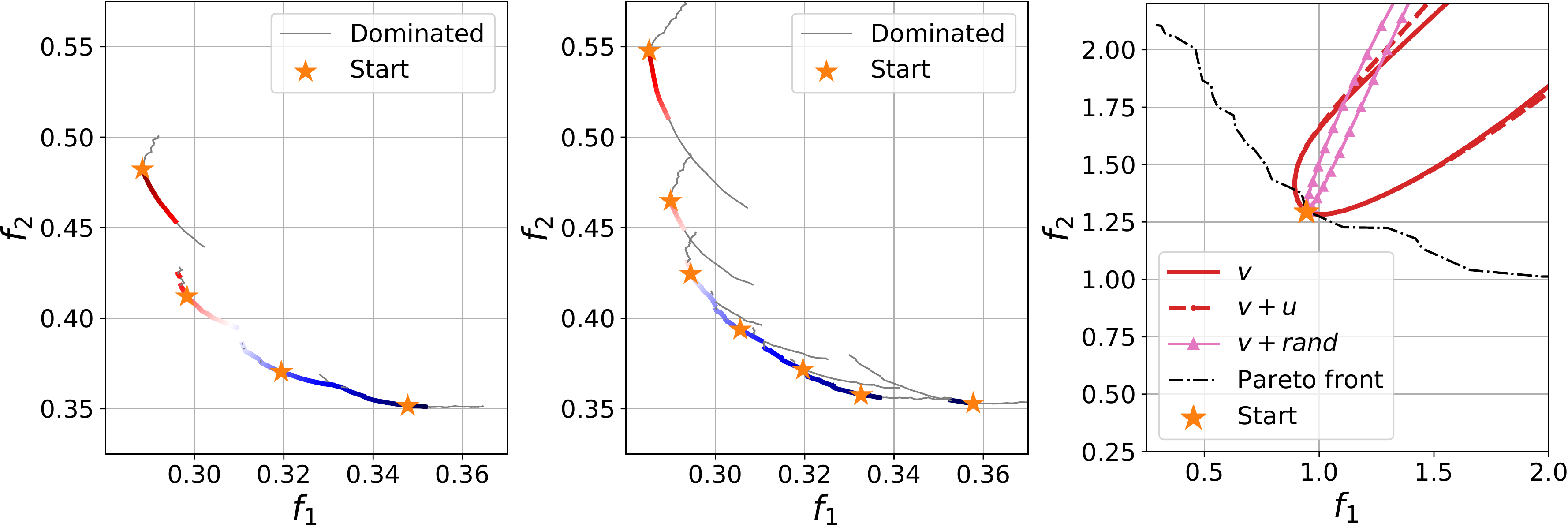}}
\caption{Illustrations of the continuous parametrization. Left: Continuous Pareto fronts grown from 4 Pareto optimal solutions (orange stars) on MultiMNIST. Curve colors indicate the value of $t$ from $-1$ (dark blue) to $1$ (dark red) and the thin gray lines indicated dominated solutions. Middle: larger approximations were formed by stitching 10 Pareto fronts. Right: comparisons between expansions with three directions: a tangent $\m{v}$ (red and solid), $\m{v}$ plus a null vector $\m{u}$ (red and dash), and $\m{v}$ plus a random direction (pink triangles).}
\label{fig:augment}
\end{center}
\vskip -0.2in
\end{figure}

From discrete solutions returned by Algorithm \ref{alg:explore}, our continuous parametrization creates low-dimensional, locally smooth Pareto sets. Moreover, we stitch them together when their Pareto fronts collide, forming a larger continuous approximation. We illustrate this idea in Figure \ref{fig:augment}: we ran Algorithm \ref{alg:explore} on MultiMNIST with $K=2$ and $N=10$ for each Pareto stationary solution $\m{x}^*$, generating two chains of solutions favoring small $f_1$ and small $f_2$ respectively. As described in Section \ref{sec:continuous}, we then constructed a piecewise linear curve parametrized by $t\in[-1,1]$. By continuously varying $t$, we explore a diverse set of solutions from favoring small $f_1$ to small $f_2$. We highlight this mapping from a single control variable to a wider-range Pareto front because it demonstrates the real advantage of a continuous reconstruction over discrete solutions. As a straightforward application, one can analyze this mapping by running single-variable gradient-descent to pick an optimal solution, which would be impossible if only discrete solutions were provided. We give more results in the supplemental material.

We conclude our discussion on continuous parametrization by demonstrating Proposition \ref{thm:null} on MultiMNIST subset in Figure \ref{fig:augment}. We precomputed its full null space and revealed over $600$ bases. We then expanded the Pareto set at a Pareto optimal $\m{x}^*$ with three directions: a tangent direction $\m{v}$, $\m{v}$ plus a null vector $\m{u}$, and $\m{v}$ plus a random direction. As expected, expanding with the first two directions led to trajectories sharing the same gradient and curvature at $\m{x}^*$, showing that we can enrich the Pareto set by adding null space bases without degrading its quality.

\begin{figure}[!t]
\vskip 0.2in
\begin{center}
\centerline{\includegraphics[width=\columnwidth]{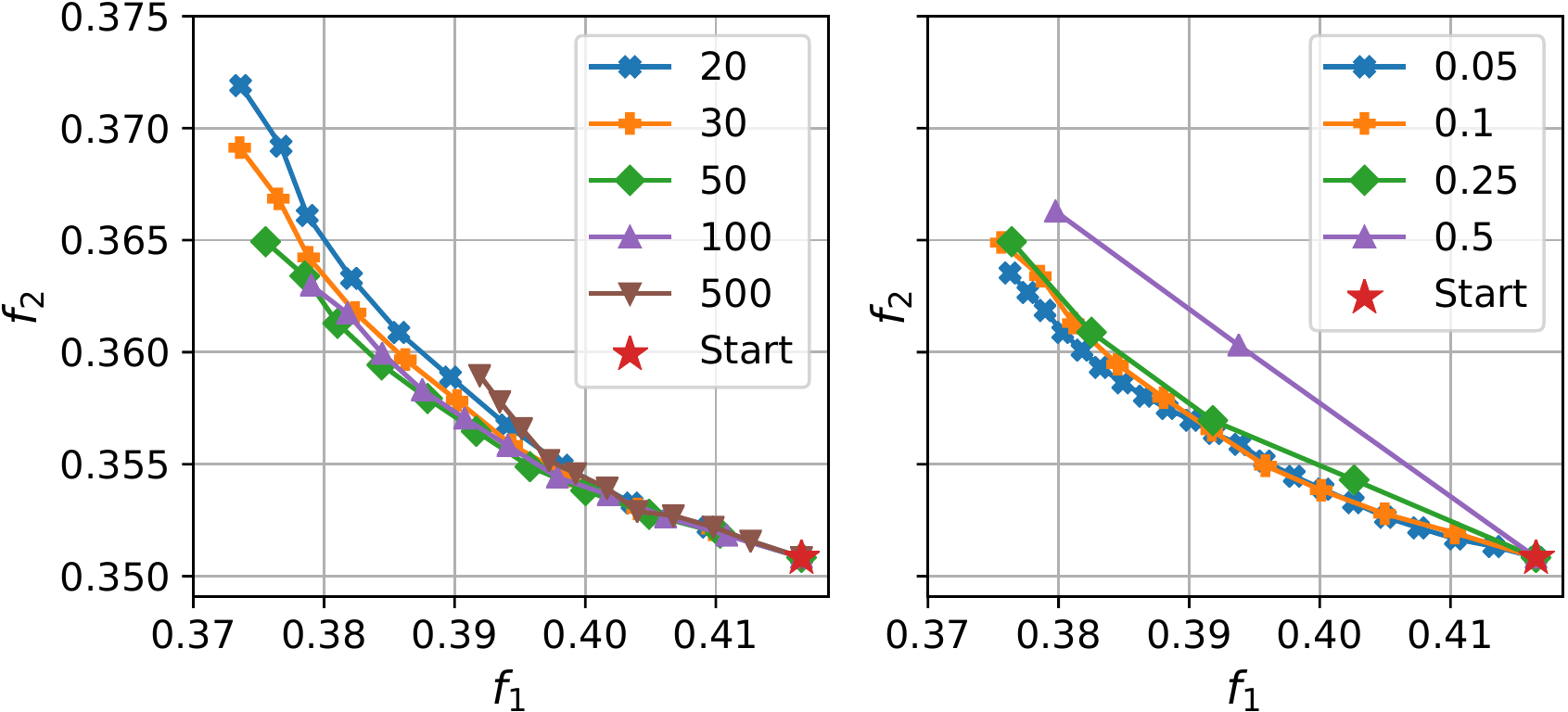}}
\caption{Pareto expansion from a Pareto optimal solution (red star) on MultiMNIST by various $k$ and $s$. Left: expansion with fixed $s$ and $k\in\{20,30,50,100,500\}$. Lower curves are more dominant. Right: expansion with fixed $k$ and $s\in\{0.05,0.1,0.25,0.5\}$. The number of runs was chosen such that its product with $s$ equals 1.}
\label{fig:ab_test}
\end{center}
\vskip -0.2in
\end{figure}

\subsection{Ablation Study}\label{sec:exp:ab_test}
Finally, we conducted ablations tests on two crucial hyperparameters in our algorithm: the maximum number of iterations $k$ in MINRES and the step size $s$ that controls the expansion speed. We started with a random Pareto stationary point $\m{x}^*$ returned by ParetoMTL, followed by running Algorithm \ref{alg:explore} with fixed parameters $K=1$ and $N=5$ on MultiMNIST and its two variants. The results are summarized in Figure \ref{fig:ab_test}, Table \ref{tb:ab_test}, and the supplemental material.

To see the influence of $k$, we fixed $s=0.1$ and ran experiments with $k\in\{20,30,50,100,500\}$, whose trajectories are in Figure $\ref{fig:ab_test}$. Between $k=20$ and $50$, the trajectories were pushed towards its lower left, indicating a better approximated Pareto front. This is as expected since more iterations in MINRES were consumed. This trend plateaued between $k=50$ and $100$. Moreover, the tail of the trajectory drifted away after $k=500$ iterations. We hypothesized that the tangent after $500$ iterations explored a new region in $\mathbb{R}^n$ where the constant step size $s=0.1$ was not proper. Based on these observations, we used $k=50$ in all experiments.

To understand how $s$ affects expansion, we reran the same experiments with a fixed $k=50$ and chose $s$ from $\{0.05,0.1,0.25,0.5\}$. For each $s$, we set the number of points to be generated to $1/s$, i.e., the product of the step size and the step number is constant. From Figure \ref{fig:ab_test} right, we noticed a conservative $s$ was likely to follow the Pareto front more closely while an aggressive step size quickly led the search to the dominated interior. This is consistent with the fact that our tangents are a first-order approximation to the true Pareto set.

\section{Conclusions}
We presented a novel, efficient method to construct continuous Pareto sets and fronts in MTL. Our method is originated from second-order analytical results in MOO, and we combined it with matrix-free iterative linear solvers to make it a practical tool for large-scale problems in MTL. We analyzed thoroughly the source of efficiency with demonstrations on synthetic examples. Moreover, experiments showed our method is scalable to modern machine learning datasets and networks with millions of parameters.

\begin{table}[tb]
\caption{Hypervolumes (HV) from the ablation study on hyperparameters $k$ and $s$. The time cost of experiments is proportional to $k$ and inverse proportional to $s$. Best results are in bold.}
\label{tb:ab_test}
\vskip 0.15in
\begin{center}
\begin{small}
\begin{sc}
\begin{tabular}{rccccc}
\toprule
$k$ & 20 & 30 & 50 & 100 & 500 \\
\midrule
hv & 7.731 & \textbf{7.739} & 7.734 & 7.727 & 7.669 \\
\midrule
$s$ & 0.05 & 0.10 & 0.25 & 0.50 & \\
\midrule
hv & \textbf{7.741} & 7.733 & 7.728 & 7.712 & \\
\bottomrule
\end{tabular}
\end{sc}
\end{small}
\end{center}
\vskip -0.1in
\end{table}

While the majority of work in MTL aims to find one near-optimal solution, we believe conflicting objectives in MTL are common and the full answer should be a wide range of candidates with varying trade-offs. Although we are not the first to explore Pareto fronts in MTL or apply second-order techniques to neural networks, we are, to our best knowledge, the first to introduce second-order analysis to Pareto exploration in MTL and the first to propose a continuous reconstruction. We believe our work enables lots of opportunities that would otherwise be impossible if only finite, sparse, and discrete solutions were given, for example, revealing the dimensionality and underlying structure of local Pareto sets, developing interpretable analysis tools for deep MTL networks, and encoding dense Pareto sets and fronts with limited storage.
\section*{Acknowledgments}

We thank Tae-Hyun Oh for his insightful suggestions and constructive feedback on Krylov subspace methods. We thank Yiping Lu for helpful discussions. We also thank all reviewers for their comments. We thank Buttercup Foshey (and Michael Foshey) for her emotional support during this work. This work is supported by the Intelligence Advanced Research Projects Activity under grant 2019-19020100001, the Defense Advanced Research Projects Agency under grant N66001-15-C-4030, and the National Science Foundation under grant CMMI-1644558.

\bibliography{main}
\bibliographystyle{icml2020}

\end{document}


\twocolumn[
\icmltitle{Efficient Continuous Pareto Exploration in Multi-Task Learning}



\icmlsetsymbol{equal}{*}

\begin{icmlauthorlist}
\icmlauthor{Pingchuan Ma}{equal,mit}
\icmlauthor{Tao Du}{equal,mit}
\icmlauthor{Wojciech Matusik}{mit}
\end{icmlauthorlist}

\icmlaffiliation{mit}{MIT CSAIL}

\icmlcorrespondingauthor{Pingchuan Ma}{pcma@csail.mit.edu}

\icmlkeywords{Multi-Objective Optimization, Multi-Task Learning, Hessian-Free Methods}

\vskip 0.3in
]



\printAffiliationsAndNotice{\icmlEqualContribution} 

\newcommand{\m}[1]{\boldsymbol{#1}}

\section{Proofs}

\subsection{Proof of Proposition 4.1}
\begin{proof}
The last constraint establishes the connection between $\m{c}$ and $\m{\alpha}$:
\begin{equation}
    \sum_{i=1}^m\alpha_i\nabla f_i(\m{x}^*_0) =(\sum_{i=1}^m\alpha_i)\m{c}=\m{c}
\end{equation}
The second equality comes from the sum of $\alpha_i$ being 1. Therefore, the optimal solution $\m{\alpha}^*$ and $\m{c}^*$ must satisfy $\m{c}^*=\nabla\m{f}(\m{x}^*_0)^\top\m{\alpha}^*$. Plugging $\m{c}^*$ back to Problem (5) reduces it to Problem (3), showing that both problems share the same optimal $\m{\alpha}^*$.
\end{proof}

\subsection{Proof of Proposition 4.2}

\begin{proof}
For simplicity, we let $\m{v}=\m{x}'(0)$. We first prove $\m{c}_{\m{v}}(t)=\m{f}(\m{x}^*+t\m{v})$ and $\m{c}_{\m{v}+u}(t)=\m{f}(\m{x}^*+t(\m{v}+\m{u}))$ have the same value and tangent direction at $t=0$, i.e., $\m{c}_{\m{v}}(0)=\m{c}_{\m{v}+\m{u}}(0)$ and $\m{c}_{\m{v}}'(0)=\m{c}_{\m{v}+\m{u}}'(0)$. The first equality is trivial because both equals $\m{f}(\m{x}^*)$. To show they have the same tangent direction, note that
\begin{equation}
\begin{aligned}
    \m{c}'_{\m{v}+\m{u}}(t)=&(f'_1(\m{x}^*+t(\m{v}+\m{u})),f'_2(\m{x}^*+t(\m{v}+\m{u})))\\
    =&\nabla \m{f}(\m{x}^*+t(\m{v}+\m{u}))(\m{v}+\m{u})
\end{aligned}
\end{equation}
Taking $t=0$ gives
\begin{equation}
    \m{c}'_{\m{v}+\m{u}}(0)=\nabla\m{f}(\m{x}^*)(\m{v}+\m{u})
\end{equation}
Since $\m{x}^*$ is Pareto optimal, we have $\m{\alpha}^\top\nabla\m{f}(\m{x}^*)=\m{0}$ (Proposition 3.1). Therefore, the dot product between $\m{\alpha}$ and $\m{c}'_{\m{v}+\m{u}}(0)$ is
\begin{equation}
    \m{\alpha}^\top\m{c}'_{\m{v}+\m{u}}(0)=\m{\alpha}^\top\nabla\m{f}(\m{x}^*)(\m{v}+\m{u})=0
\end{equation}
This indicates that $\m{c}'_{\m{v}+\m{u}}(0)$ is orthogonal to $\m{\alpha}$. Since $m=2$, we conclude $\m{c}'_{\m{v}+\m{u}}(0)$ is parallel to $(-\alpha_2,\alpha_1)$ no matter how $\m{u}$ is chosen.

We now prove the second part of the proposition. First, $\m{v}$ and $\m{u}$ are not parallel because $\m{H}(\m{x}^*)\m{u}=\m{0}$ and $\m{H}(\m{x}^*)\m{v}\neq\m{0}$. The implication is that adding $\m{u}$ to the Pareto set spanned by $\nabla f_1(\m{x}^*)$ and $\nabla f_2(\m{x}^*)$ indeed augments it by bringing a new dimension.

Second, we show $\m{c}_{\m{v}}(t)$ and $\m{c}_{\m{v}+\m{u}}(t)$ have the same curvature at $t=0$. To see this, note that the curvature of $\m{c}_{\m{d}}(t)$ at $t=0$ is defined as:
\begin{equation}
    \kappa=\frac{f'_1f''_2-f'_2f''_1}{(f'^2_1+f'^2_2)^{3/2}}
\end{equation}
where $f'_i=f'_i(\m{x}^*+t\m{d})|_{t=0}$ and $f''_i=f''_i(\m{x}^*+t\m{d})|_{t=0}$, $i=1,2$. It is now sufficient to show the denominators and numerators are the same for $\m{d}=\m{v}$ and $\m{d}=\m{v}+\m{u}$. We prove the following equality to establish the denominators are the same:
\begin{equation}
    f'_i(\m{x}^*+t\m{v})|_{t=0}=f'_i(\m{x}^*+t(\m{v}+\m{u}))|_{t=0},\quad i=1,2
\end{equation}
To see this, we expand the right-hand side:
\begin{equation}
    \begin{aligned}
        f'_i(\m{x}^*+t(\m{v}+\m{u}))|_{t=0}=&(\m{v}+\m{u})^\top\nabla f_i(\m{x}^*)\\
        =&\m{v}^\top\nabla f_i(\m{x}^*)+\m{u}^\top\nabla f_i(\m{x}^*)\\
        =&f'_i(\m{x}^*+t\m{v})|_{t=0}+\m{u}^\top\nabla f_i(\m{x}^*)
    \end{aligned}
\end{equation}
It remains to show that $\m{u}^\top\nabla f_i(\m{x}^*)=0$, or these two vectors are orthogonal. Recall that
\begin{equation}
    \alpha_1\nabla f_1(\m{x}^*)+\alpha_2\nabla f_2(\m{x}^*)=\m{0}
\end{equation}
where $\alpha_i$ comes from Proposition 3.1. Since $\alpha_1+\alpha_2=1$, at least one of them is nonzero. Without loss of generality, we assume $\alpha_1\neq0$, which gives us
\begin{equation}
    \nabla f_1(\m{x}^*)=-\frac{\alpha_2}{\alpha_1}\nabla f_2(\m{x}^*)
\end{equation}
If $\nabla f_2(\m{x}^*)=\m{0}$, $\nabla f_1(\m{x}^*)$ has to be $\m{0}$ as well, and $\m{u}^\top\nabla f_i(\m{x}_i)=0$ is trivial. Below we assume $\nabla f_2(\m{x}^*)\neq\m{0}$. Therefore, the space spanned by $\{\nabla f_1(\m{x}^*), \nabla f_2(\m{x}^*)\}$ is effectively a one-dimensional line in the direction of $\nabla f_2(\m{x}^*)$. Now consider applying Proposition 3.2 to $\m{v}$:
\begin{equation}
    \m{H}(\m{x}^*)\m{v}=\nabla f_2(\m{x}^*)\beta
\end{equation}
where $\beta$ is some scalar whose exact value is determined by Proposition 3.2. Note that the right-hand side has been simplified due to the fact that $\nabla f_1(\m{x}^*)$ and $\nabla f_2(\m{x}^*)$ are parallel. Using the fact that $\m{u}$ is a null vector of $\m{H}(\m{x}^*)$ and $\m{H}(\m{x}^*)$ is a symmetric matrix, we establish the orthogonality between $\m{u}$ and $\nabla f_2(\m{x}^*)\beta$ as follows:
\begin{equation}
    \m{u}^\top\nabla f_2(\m{x}^*)\beta=\m{u}^\top\m{H}(\m{x}^*)\m{v}=(\m{H}(\m{x}^*)\m{u})^\top\m{v}=0
\end{equation}
Since $\m{H}(\m{x}^*)\m{v}\neq\m{0}$, $\beta$ is nonzero. We then conclude $\m{u}^\top\nabla f_2(\m{x}^*)=0$. It follows that $\m{u}^\top\nabla f_1(\m{x}^*)=-\alpha_2\m{u}^\top\nabla f_2(\m{x}^*)/\alpha_1=0$.

To show the numerators are the same, we first calculate the second-order derivatives for $\m{d}=\m{v}$ as follows:
\begin{equation}
    \begin{aligned}
    f'_i(\m{x}^*+t\m{v})=&\m{v}^\top\nabla f_i(\m{x}^*+t\m{v})\\
    f''_i(\m{x}^*+t\m{v})=&\m{v}^\top\nabla^2 f_i(\m{x}^*+t\m{v})\m{v}\\
    f''_i(\m{x}^*+t\m{v})|_{t=0}=&\m{v}^\top\nabla^2f_i(\m{x}^*)\m{v}
    \end{aligned}
\end{equation}
As a result, when $\m{d}=\m{v}$, the numerator is (we simplified the notation by ignoring $\m{x}^*$ in $\nabla f_i$ and $\nabla^2 f_i$)
\begin{equation}
    \begin{aligned}
    &f'_1f''_2-f'_2f''_1\\
    =&\m{v}^\top\nabla f_1\m{v}^\top\nabla^2 f_2\m{v}-\m{v}^\top\nabla f_2\m{v}^\top\nabla^2 f_1\m{v}\\
    =&\m{v}^\top(\nabla f_1\m{v}^\top\nabla^2 f_2-\nabla f_2\m{v}^\top\nabla^2 f_1)\m{v}\\
    =&\m{v}^\top(-\frac{\alpha_2}{\alpha_1}\nabla f_2\m{v}^\top\nabla^2 f_2-\nabla f_2\m{v}^\top\nabla^2 f_1)\m{v}\\
    =&-\frac{1}{\alpha_1}\m{v}^\top\nabla f_2\m{v}^\top(\alpha_2\nabla^2f_2+\alpha_1\nabla^2f_1)\m{v}\\
    =&-\frac{1}{\alpha_1}\m{v}^\top\nabla f_2\m{v}^\top\m{H}\m{v}\\
    \end{aligned}
\end{equation}
Replacing $\m{v}$ with $\m{v}+\m{u}$ in the last equation gives us the numerator when $\m{d}=\m{v}+\m{u}$:
\begin{equation}
\begin{aligned}
&f'_1f''_2-f'_2f''_1\\
=&-\frac{1}{\alpha_1}(\m{v}+\m{u})^\top\nabla f_2(\m{v}+\m{u})^\top\m{H}(\m{v}+\m{u})\\
=&-\frac{1}{\alpha_1}\m{v}^\top\nabla f_2(\m{v}+\m{u})^\top\m{H}(\m{v}+\m{u})\\
=&-\frac{1}{\alpha_1}\m{v}^\top\nabla f_2\m{v}^\top\m{H}(\m{v}+\m{u})\\
=&-\frac{1}{\alpha_1}\m{v}^\top\nabla f_2\m{v}^\top\m{H}\m{v}
\end{aligned}
\end{equation}
where the second equality was derived with the fact $\m{u}^\top\nabla f_2=0$ and the last two equalities used $\m{H}\m{u}=\m{0}$. This shows that the two curves have the same numerators at $t=0$. Putting it together, we have proven $\m{c}_{\m{v}}(t)$ and $\m{c}_{\m{v}+\m{u}}(t)$ have the same curvature at $t=0$ when $\m{u}$ is a null vector of $\m{H}$.
\end{proof}
\section{Experimental Setup}
\subsection{ZDT2-variant}
This example has an analytic $\m{f}(\m{x}):(x_1,x_2,x_3)\in\mathbb{R}^3\rightarrow(f_1,f_2)\in\mathbb{R}^2$, defined as follows:
\begin{equation}
\begin{aligned}
    y_1=&\frac{\sin(x_1+x_2^2+x_3^2)+1}{2}\\
    y_2=&\frac{\cos(x_2^2+x_3^2)+1}{2}\\
    y_3=&y_2\\
    g=&1+\frac{9}{2}(y_2+y_3)\\
    f_1(x_1,x_2,x_3)=&y_1\\
    f_2(x_1,x_2,x_3)=&g-\frac{y_1^2}{g}
\end{aligned}
\end{equation}
The Pareto front is given by
\begin{equation}
f_2=1-f_1^2\quad f_1\in[0,1]    
\end{equation}
and the analytic Pareto set is
\begin{equation}
    x_2^2+x_3^2=(2k+1)\pi,\quad k=0,1,2,\cdots
\end{equation}
which is a family of concentric cylinders. In the paper, we analyzed the innermost Pareto set $x_2^2+x_3^2=\pi$. The rightmost figure in Figure 2 in the paper was generated by plotting $\m{f}(\m{x}^*+s\m{d}), s\in[-0.1,0.1]$ with $\m{d}$ being a unit vector of $\nabla f_1(\m{x}^*)$, $\nabla f_2(\m{x}^*)$, and the approximated tangent directions after $2$ iteration of MINRES respectively.

The experiments in Figure 3 of the main paper were set up as follows: starting with a randomly chosen Pareto optimal $\m{x}^*$, we spawned a new $\m{x}$ by computing $\m{x}=\m{x}^*+0.1\m{d}$ where $\m{d}$ is a unit vector calculated from two methods: 1) running MINRES for $2$ iteration to get the approximated tangent direction; 2) perturbing $\alpha$ at $\m{x}^*$ to get $\alpha'$ and letting $\m{d}=\alpha'_1(\m{x}^*)+\alpha'_2\nabla f_2(\m{x}^*)$. The second method is the WeightedSum baseline introduced in the main paper and can be interpreted as exploring by running one iteration of gradient-descent to minimize $\alpha'_1f_1+\alpha'_2f_2$. We then used MGDA \cite{desideri12} plus line search to push new $\m{x}$ back to the Pareto front. The step size in our line search was initially $1$ and decayed by $0.9$ exponentially.

\subsection{MultiMNIST Subset}
We first generated the full MultiMNIST dataset (see Section \ref{sup:sec:mnist}) and picked a subset of $2048$ images, downsampled from $28\times28$ to $14\times14$, as our MultiMNIST Subset example. The two objectives are the cross entropy losses of classifying the top-left and bottom-right digits evaluated on all $2048$ images. Regarding the classifier, we used a modified LeNet5 \cite{lecun98} network, which has $1500$ parameters. Our modified network starts with a convolutional layer with 10 channels, a $5\times5$ kernel, and a stride of 2 pixels, followed by a $2\times2$ max pooling layer. Next, the results are fed into a fully connected layer of size $20\times10$ and then sent to two fully connected layers, one for each task. We use ReLU as the nonlinear function in the network. Essentially, this synthetic example attempts to use a small network to overfit $2048$ images. To generate the Pareto front, we ran BFGS \cite{nocedal06} to optimize $w_1f_1+w_2f_2$ with $w_1=0,0.01,0.02,\cdots,1$ from the same random initial guess, which generated a list of 101 solutions $\m{x}^*_0,\m{x}^*_1,\m{x}^*_2,\cdots,\m{x}^*_{101}$. We then linearly interpolated $\m{f}(\m{x}^*_i),i=0,1,2,\cdots,100$ and treat the resulting piecewise linear spline as the (empirical) Pareto front.

The experiment in Figure 4 of the main paper was conducted as follows: starting with a randomly chosen $\m{x}^*_i$, we plotted $\m{f}(\m{x}^*_i+s\m{d}),s\in[-0.5,0.5]$ where $\m{d}$ is a unit vector of the approximated tangent direction. We got the tangent direction by running $50$ MINRES iterations to solve Equation (6) of the main paper with $\m{\beta}$ sampled from a standard normal distribution. In particular, we found gradient correction (Equation (5) of the main paper) useful in this example. We then ran 50 iterations of gradient-descent (GD) to minimize $f_1$, $f_2$, and $w_1f_1+w_2f_2$ respectively. Here $w_1$ and $w_2$ are perturbed from the corresponding $\m{\alpha}$ vector at $\m{x}^*_i$. This shows how well gradient-descent can explore the Pareto front within the time budget of $50$ times of back-propagation. We used $1/\sqrt{t+1}$ where $t$ is the iteration index to decay the learning rate in GD from $0.005$.

\subsection{MultiMNIST and Its Variants}\label{sup:sec:mnist}
\begin{figure}[!t]
\vskip 0.2in
\begin{center}
\centerline{\includegraphics[width=\columnwidth]{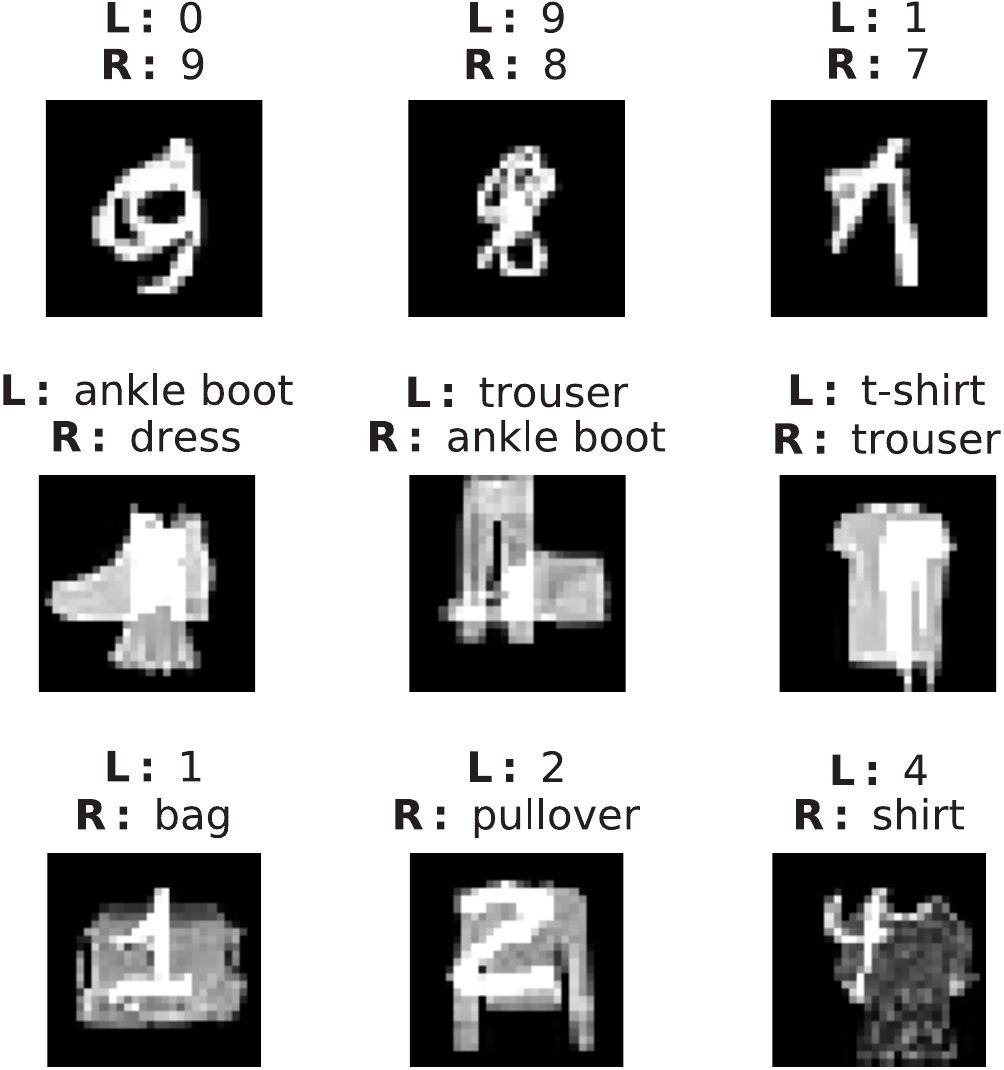}}
\caption{Sample images from MultiMNIST (top), MultiFashion (middle), and MultiFashionMNIST (bottom). Above each image are the labels of the upper-left (L) and lower-right (R) items.}
\label{fig:sup:mnist}
\end{center}
\vskip -0.2in
\end{figure}

\paragraph{Dataset and Task Description} We followed Sabour et al. \yrcite{sabour17} to generate MultiMNIST, FashionMNIST, and MultiFashionMNIST. We first created $36\times36$ images by placing two $28\times28$ images from MNIST or FashionMNIST \cite{xiao17} in the upper-left and lower-right corner with a random shift of up to 2 pixels in each direction. The synthesized images were then resized to 28$\times$28 and normalized with a mean of 0.1307 and a standard deviation of 0.3081. No data augmentation was used for training or testing. Following ParetoMTL \cite{lin19}, we built MultiMNIST from MNIST, MultiFashion from FashionMNIST, and MultiFashionMNIST from both (Figure \ref{fig:sup:mnist}). Each dataset has 60,000 training images and 10,000 test images. The objectives are the cross entropy losses of classifying the upper-left and lower right items in the image.

\paragraph{Network Architecture} The backbone network is a modified LeNet \cite{lecun98}. Our network starts from two convolutional layers with a $5\times5$ kernel and a stride of 1 pixel. The two layers have 10 and 20 channels respectively. A fully connected layer of 50 channels appends the convolutional layers, which is then followed by two 10-channel fully connected layers, one for each task. We add a $2\times2$ max pooling layer right after each convolutional layer and use ReLU as the nonlinear function. The network contains 22,350 trainable parameters.

\paragraph{Training} We trained all baselines for 30 epochs of SGD. We used 256 as our mini-batch size and set the momentum to 0.9. The learning rate started from 0.01 and decayed with a cosine annealing scheduler.

For our method, we used 50 iterations of MINRES to solve Equation (4) of the main paper with the right-hand side sampled between $\nabla f_1(\m{x}^*_0)$ and $\nabla f_2(\m{x}^*_0)$. We did not correct the gradients (Equation (5) of the main paper) in this example as we found using the original gradients were more effective.

\subsection{UCI Census-Income}\label{sec:sup:uci_setup}
\paragraph{Dataset and Task Description} UCI Census-Income \cite{kohavi96} is a demographic dataset consisting of information about around 300,000 adults in the United States. Lin et al. \yrcite{lin19}, one of the state-of-the-art baselines, proposed three tasks on this dataset: 1) whether the person's income exceeds \$50K/year, 2) whether the person's education level is at least college, and 3) whether the person is never married. We did not use their first task because the results are highly imbalanced ($93.8\%$ of the dataset would have the same label). Instead, our first task is whether the person's age is greater than or equal to $40$. The tasks were evaluated by cross-entropy losses. We converted all categorical data into one-hot vectors and concatenated them along with continuous data into a $487$ dimensional feature vector. After removing invalid data, training and test sets have 199,523 and 99,762 samples respectively.

\paragraph{Network Architecture} We used a multilayer perceptron (MLP) with two hidden layers of 256 and 128 channels as the shared feature extractor and a fully connected layer as the classifier for each task. We chose ReLU as the nonlinear activation function. This network contains 158,598 trainable parameters in total.

\paragraph{Training} We trained all baselines with 30 epochs of SGD and used a mini-batch of size 256 and a momentum of 0.9. The learning rate started from 0.001 and decayed with a cosine annealing scheduler.

For our method, we used 100 iterations of MINRES to solve the tangent direction and gradient correction was not used. The right-hand side of Equation (4) was sampled as follows: for each task $f_i$, we flipped a coin to determine a binary label $l_i\in\{0,1\}$. The right-hand side was then the sum of all $\nabla f_i(\m{x}^*_0)$ with $l_i=1$. We skipped a sample if $l_1=l_2=l_3$.

\subsection{UTKFace}
\paragraph{Dataset and Task Description} UTKFace \cite{zhang17} is a dataset of over 20,000 face images. Each image has $200\times200$ pixels and 3 color channels. We considered three tasks on this dataset: 1) predicting the age of each face, 2) classifying the gender, and 3) classifying the race. We used the Huber loss with $\delta=1$ for task 1 and cross entropy losses for task 2 and 3. We preprocessed the age information by normalizing it to the standard normal distribution. Moreover, each image was resized to $64\times64$ and each pixel was further normalized with mean values and standard deviations from ImageNet \cite{deng09}. We created the training and test set with an 80/20 split of UTKFace. After data cleaning, our training and test sets have 18,964 and 4,741 images respectively.

\paragraph{Network Architecture} Our network was built upon a standard ResNet18 \cite{he16} by appending a fully connected layer to it for each task. Batch normalization \cite{ioffe15} was used with a momentum of 0.1. The network contains 11,180,616 trainable parameters.

\paragraph{Training} We ran all baseline experiments with 30 epochs of SGD and a mini-batch size 256. We used a weight decay of $1\mathrm{e}{-5}$ and a momentum of 0.9. The learning rate started from 0.01 and decayed with a cosine annealing scheduler. Batch normalization was frozen when we were expanding the Pareto front from a Pareto optimal network.

For our method, the training process was the same as in UCI Census-Income except that we used $50$ MINRES iterations instead of $100$.
\section{Synthetic Examples}

\subsection{ZDT2-variant}
\begin{figure}[!h]
\vskip 0.2in
\begin{center}
\centerline{\includegraphics[width=\columnwidth]{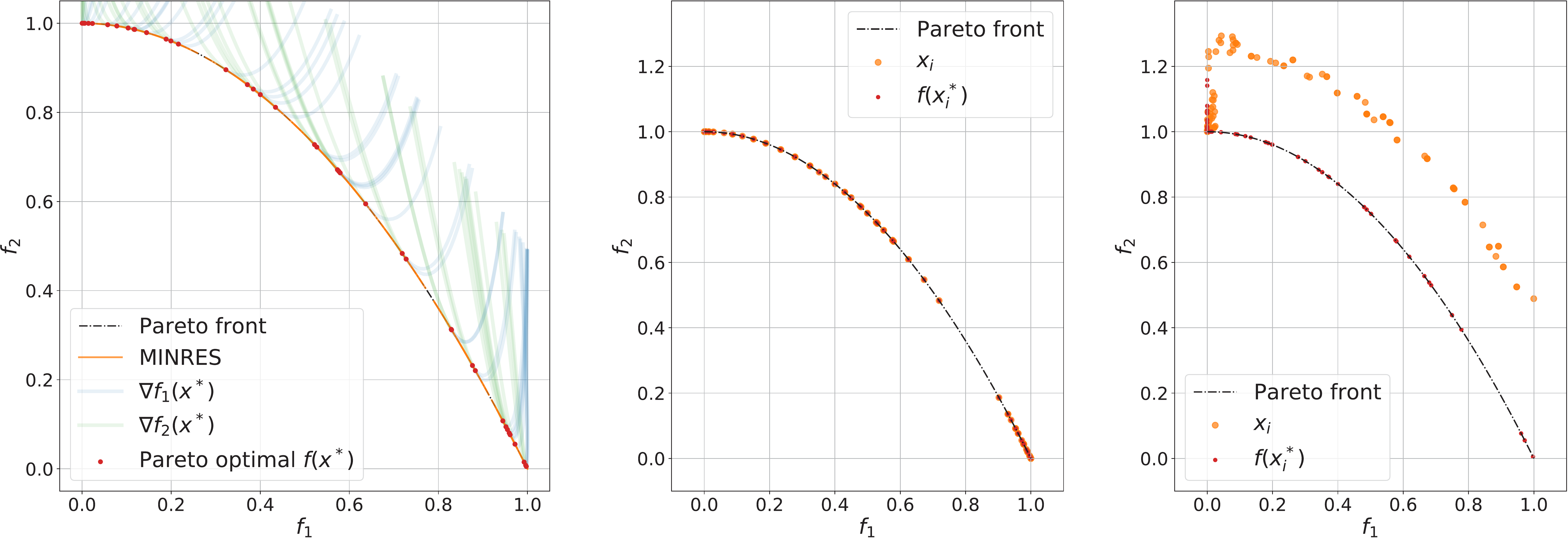}}
\caption{More experimental results on ZDT2-variant. Left: plotting $\m{f}(\m{x}^*+s\m{d}),s\in[-0.1,0.1]$ with 40 random $\m{x}^*$ (red) and $\m{d}$ being tangent directions (orange) and gradients (blue and green); Middle and right: running Algorithm 1 with MGDA as the optimizer and comparing two expansion strategies: moving along the tangent directions from MINRES after 2 iterations (middle) and walking along the perturbed weighted sum of gradients (right). The experiments were repeated on 10 random seeds, and all explored points on the Pareto front are colored in red. Results returned by \texttt{ParetoExpand} are colored in orange.}
\label{fig:sup:zdt2_variant}
\end{center}
\vskip -0.2in
\end{figure}

Here we present more experimental results on ZDT2-variant from multiple random seeds. Figure \ref{fig:sup:zdt2_variant} left shows 40 random Pareto optimal solutions and expansions from them with tangent directions and gradients. This essentially repeated Figure 2 in the main paper 40 times. It can be seen that tangent directions behave consistently better than gradients in terms of exploring the Pareto front across all random samples. Figure \ref{fig:sup:zdt2_variant} middle and right implemented Algorithm 1 from 10 random seeds and collected 10 Pareto optimal solutions each time. This experiments duplicated Figure 3 in the main paper with 10 different random seeds, and they show that using tangent directions allows us to slide on the Pareto front closely as expected. It is worth noting that part of the solutions optimized by MGDA clustered along the line segment $f_1=0,f_2\geq1$. Due to the design of ZDT2-variant, solutions like these are not Pareto optimal but Pareto stationary, and thus MGDA could not make further progress from them. Moreover, we report the time cost in Table \ref{tb:sup:zdt2_variant}, which confirms that the time saving mostly comes from the near-optimal inputs to \texttt{ParetoOptimize}.

\begin{table}[!b]
\caption{The number of evaluations of objectives ($\m{f}$), gradients ($\nabla\m{f}$), and Hessian-vector products ($\nabla^2\m{f}$) in Figure \ref{fig:sup:zdt2_variant} middle (ours) and right (WeightedSum). The abbreviation EXP and OPT means the time cost from \texttt{ParetoExpand} and \texttt{ParetoOptimize} respectively.}
\label{tb:sup:zdt2_variant}
\vskip 0.15in
\begin{center}
\begin{small}
\begin{sc}
\begin{tabular}{rccc}
\toprule
Method & $\#\m{f}$ & $\#\nabla\m{f}$ & $\#\nabla^2\m{f}$ \\
\midrule
Ours (Exp) & 0 & 50 & 300 \\
Ours (Opt) & 100 & 100 & 0 \\
\midrule
WeightedSum (Exp) & 0 & 50 & 0 \\
WeightedSum (Opt) & 17931 & 1818 & 0\\
\bottomrule
\end{tabular}
\end{sc}
\end{small}
\end{center}
\vskip -0.1in
\end{table}

\subsection{MultiMNIST Subset}
\begin{figure}[!t]
\vskip 0.2in
\begin{center}
\centerline{\includegraphics[width=\columnwidth]{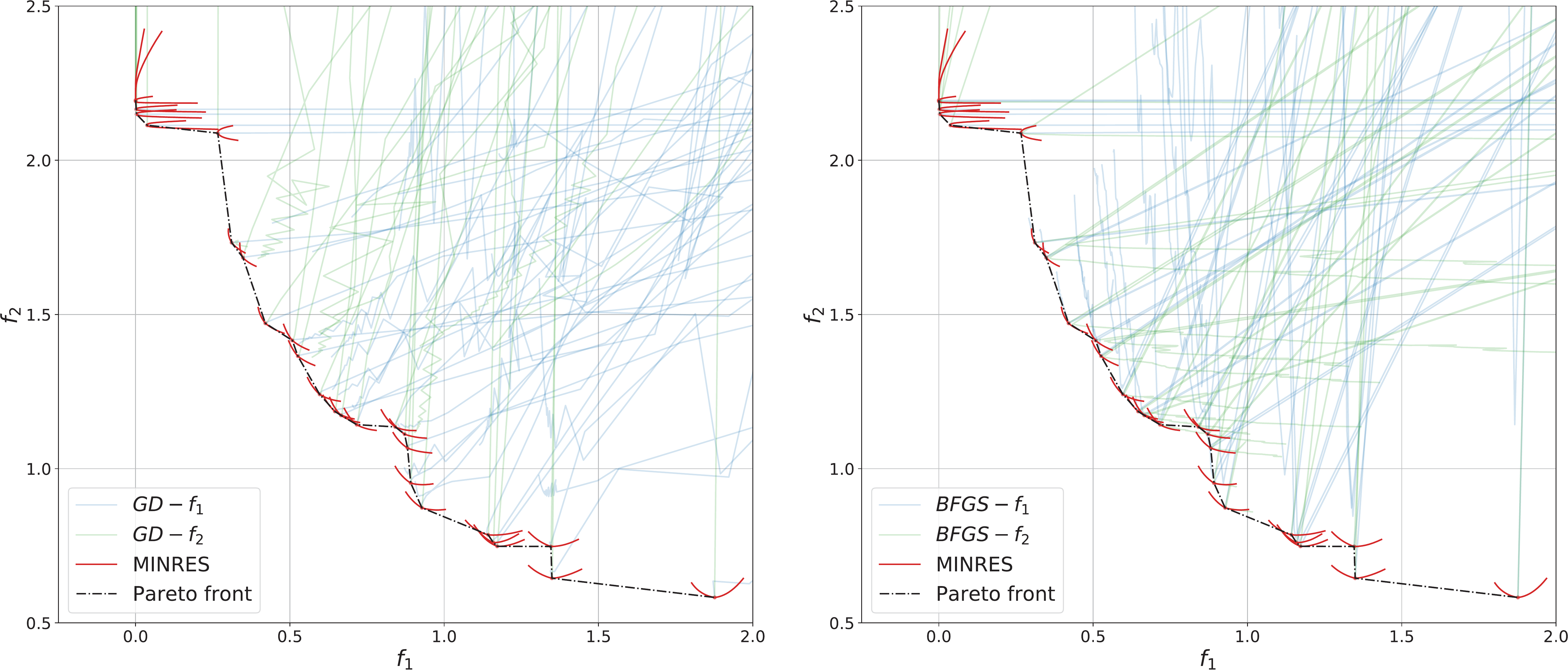}}
\caption{Comparisons of three Pareto expansion strategies on MultiMNIST Subset. The empirical Pareto front is plotted in black with Pareto optimal solutions $\m{x}^*$ drawn as 26 red dots. The red curves in both figures show $\m{f}(\m{x}^*+s\m{v}),s\in[-0.5,0.5]$ where $\m{v}$ is the tangent from $50$ iterations of MINRES. We used $50$ iterations of GD (left) and BFGS (right) to minimize $f_1$ and $f_2$ from $\m{x}^*$, with intermediate solutions shown in blue and green respectively.}
\label{fig:sup:m3_expand}
\end{center}
\vskip -0.2in
\end{figure}

We now present more results on MultiMNIST Subset in Figure \ref{fig:sup:m3_expand} as we extended the experiments in Figure 4 of the main paper. We sampled 26 Pareto optimal points $\{\m{x}^*_i\}$ evenly distributed on the empirical Pareto front. For each of them, we depicted $\m{f}(\m{x}^*_i + s\m{v}),s\in[-0.5,0.5]$ (red) where $\m{v}$ is returned by running MINRES after $50$ iterations. Furthermore, we minimized $f_1$ and $f_2$ from $\m{x}^*_i$ with $50$ iterations of GD and BFGS and plotted the trajectory of intermediate solutions at each iteration (blue and green). It can be seen from Figure \ref{fig:sup:m3_expand} that the tangent directions expanded the empirical Pareto front more accurately and clearly dominated the region explored by GD or BFGS.

\section{Pareto Expansion}
In this section, we repeated the two experiments described in Section 6.3 of the main paper on all five datasets with more random seeds. Essentially, the results in this sections extend Figure 5 and Figure 6 of the main paper. To recap, the first experiment uses our Pareto expansion method to grow dense Pareto fronts from known Pareto optimal solutions, and the second experiment compares our method to the WeightedSum baseline to establish the necessity of using tangent directions. For simplicity, we will call them sufficiency and necessity experiments respectively.

\subsection{MultiMNIST and Its Variants}
\begin{figure*}[!tb]
\vskip 0.2in
\begin{center}
\centerline{\includegraphics[width=\textwidth]{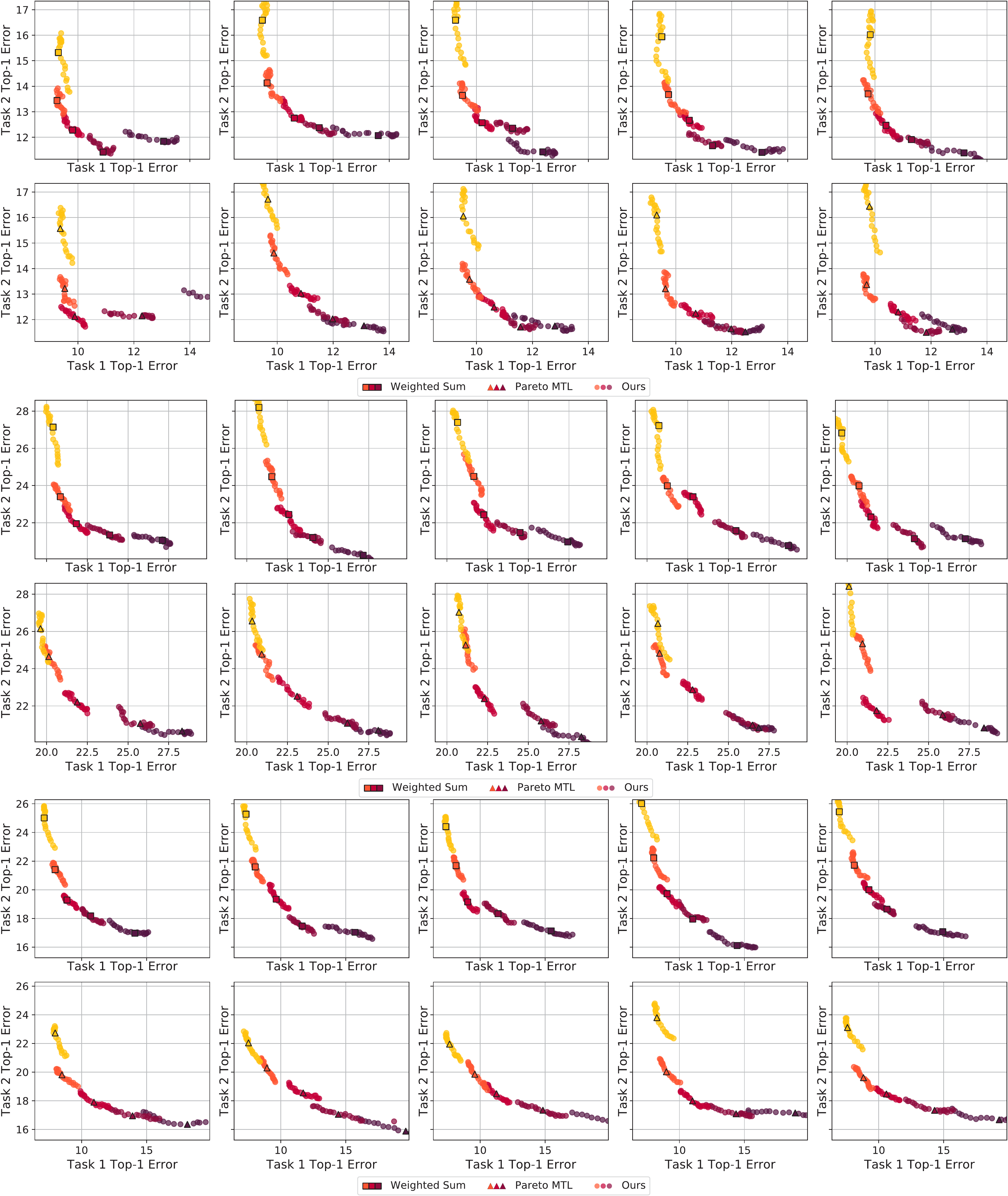}}
\caption{Expanding the Pareto front with our method on MultiMNIST (top two rows), MultiFashion (middle two rows), and MultiFashionMNIST (bottom two rows) from 5 Pareto optimal seeds generated by the WeightedSum baseline (squares) and ParetoMTL (triangles) with different initial random guesses (left to right). Our method grew dense Pareto fronts (colorful circles) from these 5 seeds.}
\label{fig:sup:multi:expand}
\end{center}
\vskip -0.2in
\end{figure*}

Figure \ref{fig:sup:multi:expand} displays the results of our sufficiency experiment on MultiMNIST and its two variants. We grew Pareto fronts from 5 seeds optimized by two baselines: WeightedSum and ParetoMTL. This figure is an extension to Figure 5 in the main paper. We stress again that growing such dense Pareto fronts only took a fraction of the training time spent on getting one Pareto optimal solution from baselines.

\begin{figure*}[!tb]
\vskip 0.2in
\begin{center}
\centerline{\includegraphics[width=\textwidth]{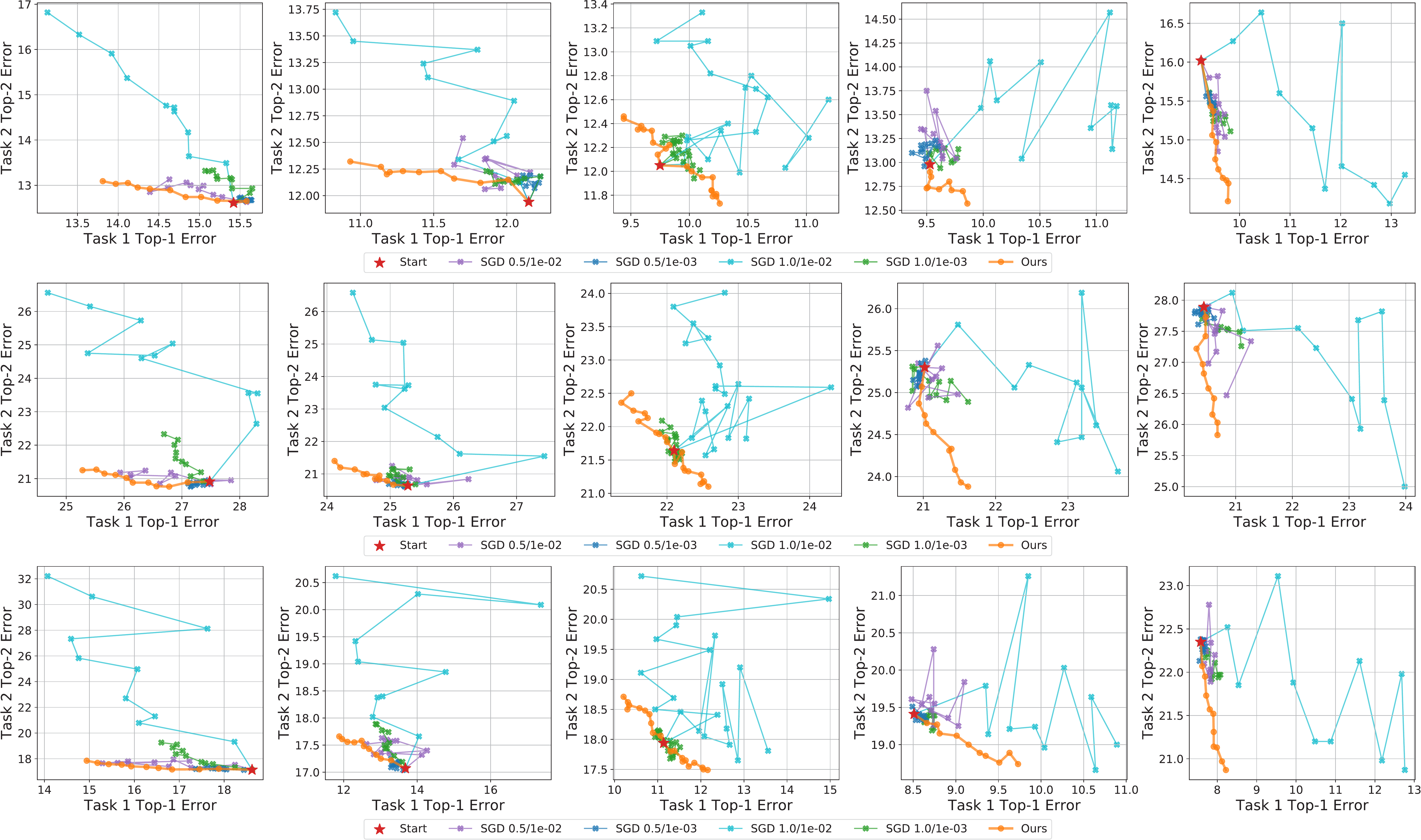}}
\caption{Comparisons of two expansion methods (ours and running SGD with a weighted sum) from a given Pareto optimal solution (red star). Top to bottom: results on MultiMNIST, MultiFashion, and MultiFashionMNIST. Left to right: we started the experiments from five different Pareto optimal solutions found by ParetoMTL. In all figures, lower left means better solutions. All SGD methods are labeled with preference on task 1/learning rate.}
\label{fig:sup:multi:fine_tune}
\end{center}
\vskip -0.2in
\end{figure*}

Similarly, we reran the necessity experiment and summarized the results in Figure \ref{fig:sup:multi:fine_tune}. For each dataset, we repeated the experiment on 5 different Pareto optimal solutions found by ParetoMTL (squares and triangles in Figure \ref{fig:sup:multi:expand}). We second that in all figures, lower left indicates better performances, and the region expanded by our method (orange lines) dominates SGD with various learning rates and weight combinations.

\subsection{UCI Census-Income}
\begin{figure*}[!tb]
\vskip 0.2in
\begin{center}
\centerline{\includegraphics[width=\textwidth]{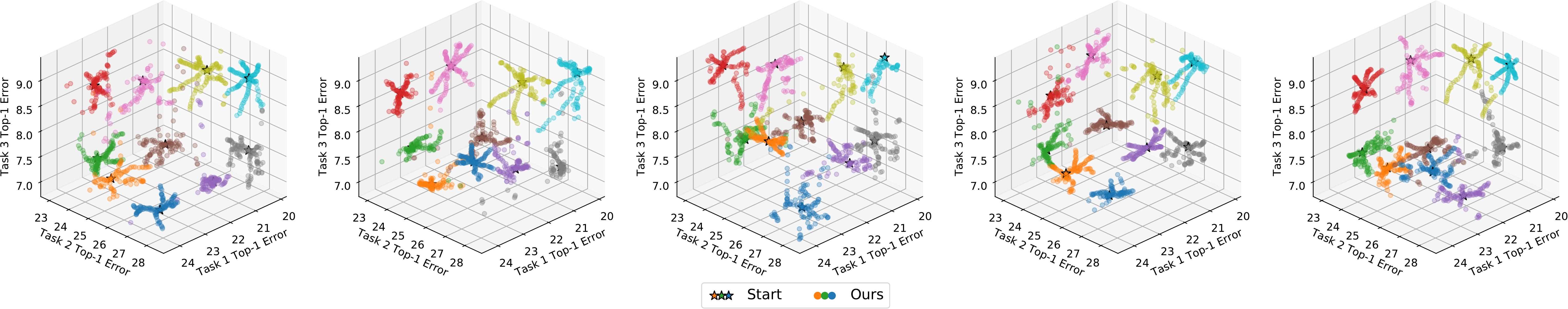}}
\caption{Expanding the Pareto front with our method on UCI Census-Income from 10 Pareto optimal seeds generated by the WeightedSum baseline. Five random initial guesses (left to right) were used to generate these results.}
\label{fig:sup:uci:expand}
\end{center}
\vskip -0.2in
\end{figure*}

Figure \ref{fig:sup:uci:expand} displays the result of the sufficiency experiment. Note that this dataset has three objectives. We repeated this experiment with 5 random seeds. For each random seed, we ran SGD 10 times with different weight combinations to generate 10 Pareto optimal solutions that are evenly distributed on the Pareto front, which is roughly a concave surface viewed from the camera position. Points with smaller values (farther away from the camera in the figure) are preferred.

\begin{figure*}[!tb]
\vskip 0.2in
\begin{center}
\centerline{\includegraphics[width=\textwidth]{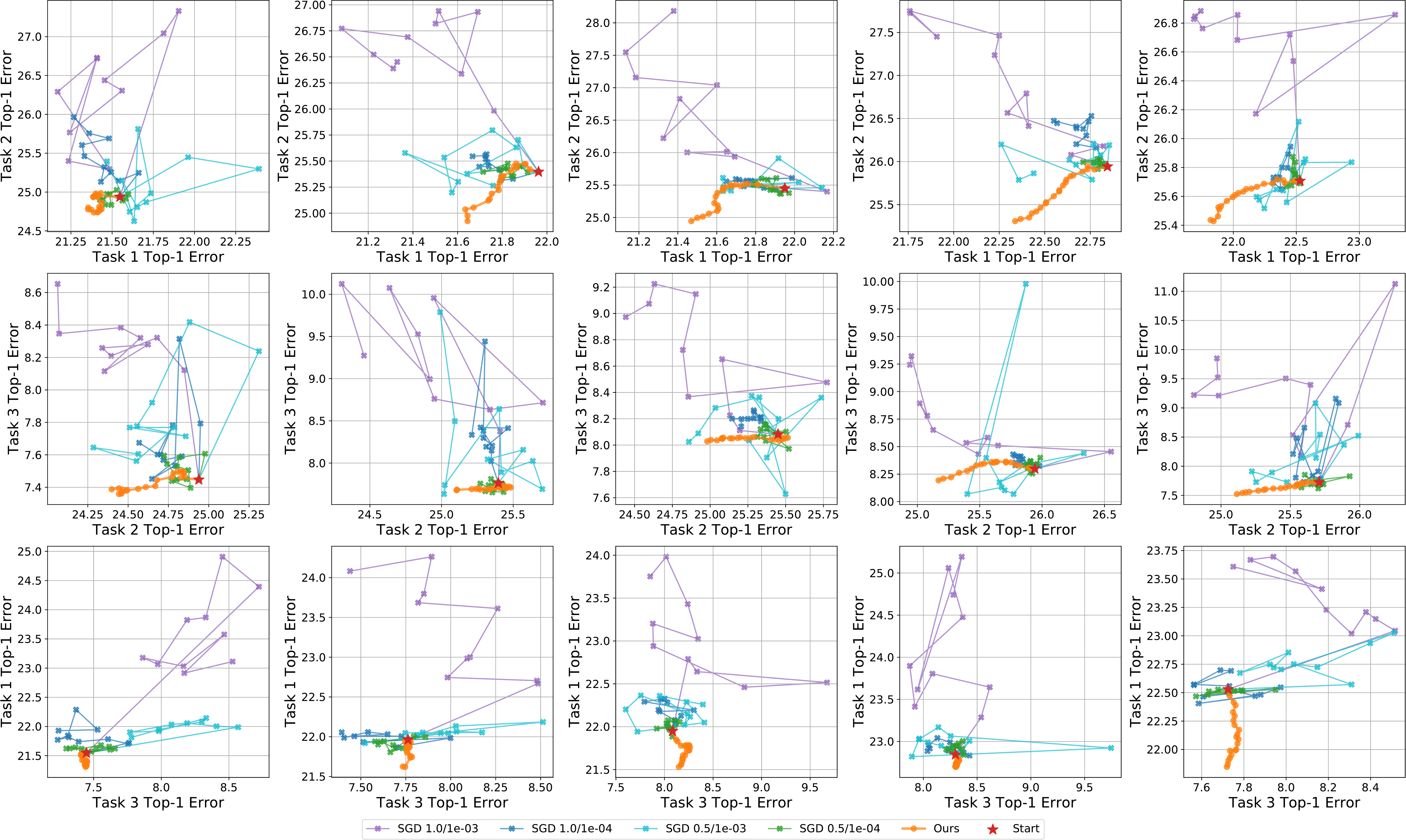}}
\caption{Comparisons of two expansion methods (ours and running SGD with a weighted sum) from a given Pareto optimal solution (red star) on UCI Census-Income. Left to right: we started the experiments from five different Pareto optimal solutions found by SGD with weights $(1/3,1/3,1/3)$. In all figures, lower left means better solutions. All SGD methods are labeled with preference on task of the horizontal axis/learning rate.}
\label{fig:sup:uci:fine_tune}
\end{center}
\vskip -0.2in
\end{figure*}

Furthermore, Figure \ref{fig:sup:uci:fine_tune} summarizes the necessity experiment on this dataset. We first ran SGD to minimize a combination of three objectives with a preference vector $(1/3,1/3,1/3)$ to obtain a Pareto optimal solution $\m{x}^*$. We then considered three pairs of losses $(f_i,f_j)$ where $(i,j)\in\{(1,2),(2,3),(3,1)\}$. For each $(f_i,f_j)$ pair, we ran MINRES from $\m{x}^*$ and compared it to SGD baselines with different weight combinations and learning rates. For all figures, lower left region is Pareto optimal. In most cases, Pareto fronts revealed by our method dominate SGD results.

\subsection{UTKFace}
\begin{figure*}[!tb]
\vskip 0.2in
\begin{center}
\centerline{\includegraphics[width=\textwidth]{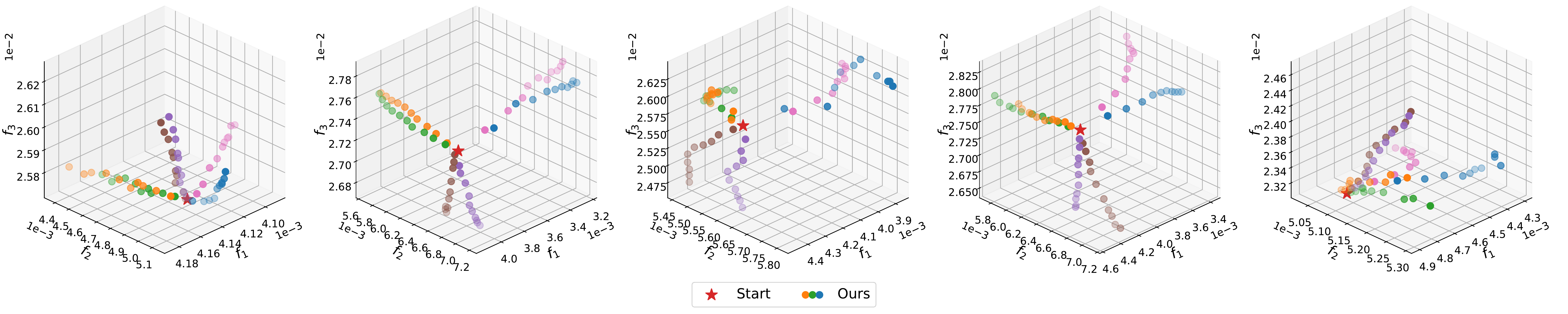}}
\caption{Expanding the Pareto front with our method on UTKFace from five random initializations. We grew our solutions from a seed (red star) to 6 directions (colorful circles) computed by 50 MINRES iterations.}
\label{fig:sup:face:expand}
\end{center}
\vskip -0.2in
\end{figure*}

The sufficiency experiment is reported in Figure \ref{fig:sup:face:expand}. We randomly picked 5 initial networks and ran SGD to minimize a combination of three objectives with a weight vector $(1/3,1/3,1/3)$. We then expanded the local Pareto front with our method by running 50 MINRES iterations 6 times, generating 6 trajectories from the Pareto optimal solution. The choice of $6$ comes from the fact that three objectives have 8 possible combinations of binary labels (see Section \ref{sec:sup:uci_setup}) and we skipped combinations of all-zero or all-one labels in the sufficiency experiment.

\begin{figure*}[!tb]
\vskip 0.2in
\begin{center}
\centerline{\includegraphics[width=\textwidth]{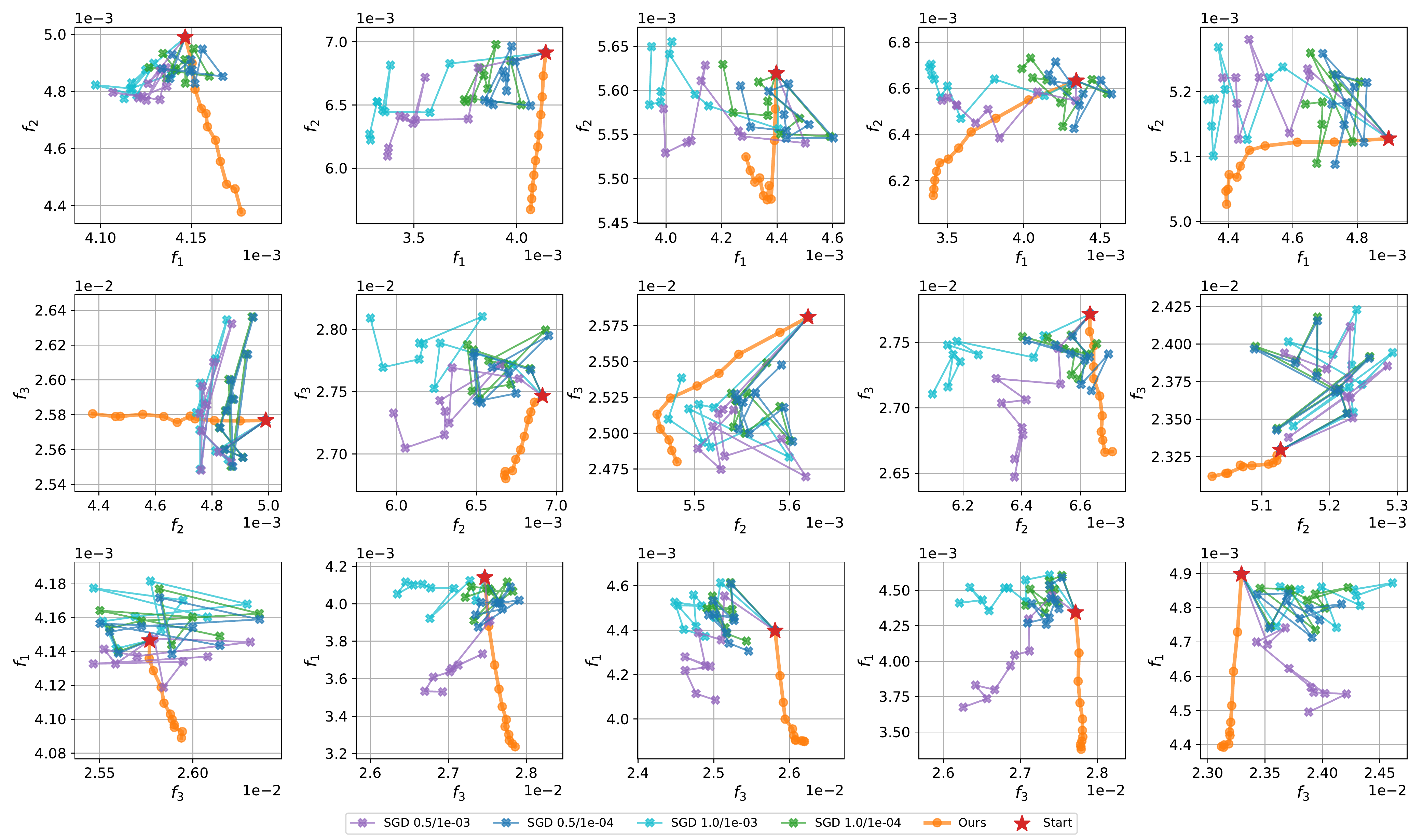}}
\caption{Comparisons of two expansion methods (ours and running SGD with a weighted sum) from a given Pareto optimal solution (red star) on UTKFace. Left to right: we started the experiments from five different Pareto optimal solutions found by SGD with weights $(1/3,1/3,1/3)$. In all figures, lower left means better solutions. All SGD methods are labeled with preference on task of the horizontal axis/learning rate.}
\label{fig:sup:face:fine_tune}
\end{center}
\vskip -0.2in
\end{figure*}

We present the necessity experiment in Figure \ref{fig:sup:face:fine_tune}. Since both UTKFace and UCI Census-Income have three objectives, we inherited the same experiment setup from UCI Census-Income. Methods that can explore towards the lower left region are preferred. Among the 15 experiments and 4 SGD baselines reported in Figure \ref{fig:sup:face:fine_tune}, we summarize that our method almost dominated all SGD baselines in 5 experiments (row 1 column 4, row 2 column 3, and the rightmost column), was clearly outperformed by one SGD baseline in our experiment (purple in row 2 column 4), and performed comparably in the remaining experiments.

\section{Continuous Parametrization}
In this section, we present results that extend Figure 8 of the main paper. The main idea we want to demonstrate is twofold: locally, we show that Pareto optimal solutions found by our method can be used as backbones to grow a continuous, approximated Pareto front; Globally, such Pareto fronts can be stitched together to cover a wide range of solutions with varying trade-offs.

\subsection{MultiMNIST and Its Variants}
\begin{figure*}[!tb]
\vskip 0.2in
\begin{center}
\centerline{\includegraphics[width=\textwidth]{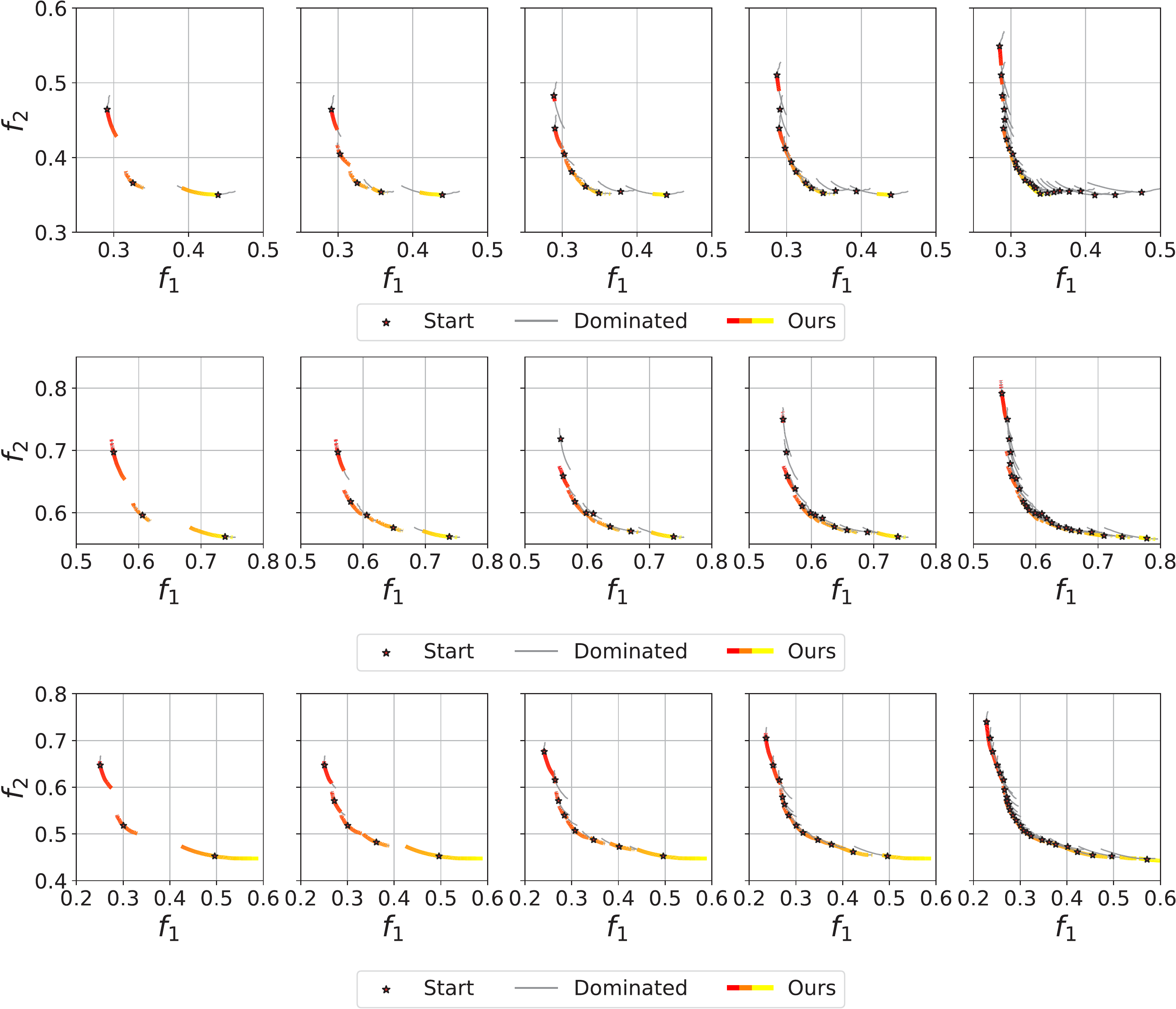}}
\caption{Continuous parametrization on MultiMNIST (top), MultiFashion (middle), and MultiFashionMNIST (bottom). From left to right: we gradually increased the number of Pareto optimal solutions (red stars) obtained from running SGD with different weights. We then ran Algorithm 1 to grow a continuous Pareto front from each solution (colorful circles) and filtered out dominated solutions (gray). The red-to-yellow color indicates the value of the scalar parameter that traverses the whole final Pareto front.}
\label{fig:sup:mnist_continuous}
\end{center}
\vskip -0.2in
\end{figure*}

Figure \ref{fig:sup:mnist_continuous} depicts the continuous parametrization on MultiMNIST and its two variants. For each dataset, we gradually increased the number of Pareto optimal seeds from 3 to 25 and reconstructed a continuous approximation of the Pareto front (a curve in this 2D case) from each seed. It can be seen that as we added more seeds, the continuous Pareto fronts became more connected. By stitching them together, we have created a union of continuous Pareto fronts that offers very diverse choices of trade-offs. We further reparametrized it with a single scalar for easy manipulation and intuitive visualization.

\subsection{UCI Census-Income}

\begin{figure*}[!tb]
\vskip 0.2in
\begin{center}
\centerline{\includegraphics[width=\textwidth]{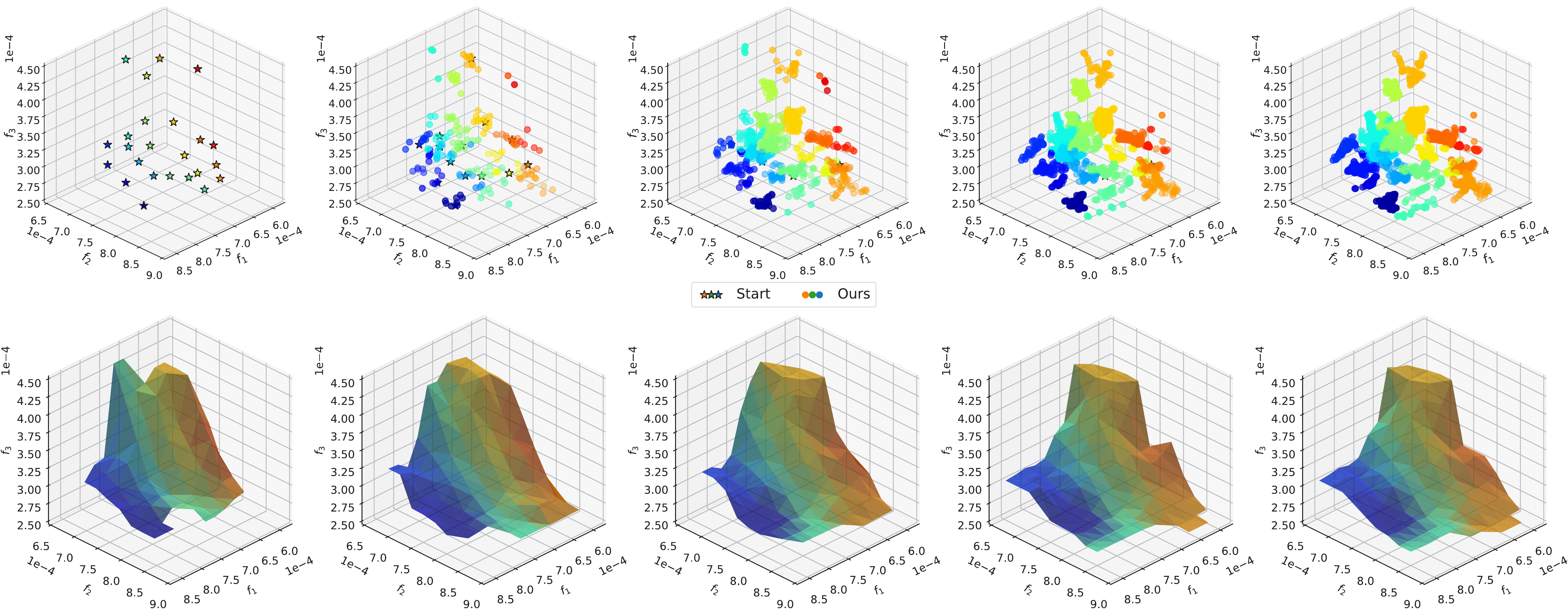}}
\caption{Continuous parametrization on UCI Census-Income. Top: starting with discrete Pareto optimal seeds returned by running SGD with various weights on objectives (colorful stars), we constructed the continuous Pareto sets (not shown) with our method, densely sampled new solutions from these Pareto sets, and plotted their performances (colorful circles). Samples from the same seed share the same color. Bottom: we reconstructed a continuous surface mesh to approximate the Pareto front revealed by these samples above. The color on the surface mesh has a one-to-one correspondence to the color of Pareto optimal seeds; Left to right: We gradually increased the number of samples to show the progress of our reconstruction.}
\label{fig:sup:uci_continuous}
\end{center}
\vskip -0.2in
\end{figure*}

We display the continuous parametrization results on UCI Census-Income in Figure \ref{fig:sup:uci_continuous}. We started with 36 Pareto optimal seeds, densely sampled the continuous Pareto set grown from each seed to evaluate their performances, and labeled samples from the same patch with a unique color. We gradually increased the number of samples in order to show how our continuous Pareto fronts were constructed progressively. Additionally, we reconstructed a 3D surface mesh from the Pareto fronts for better visualization.

\subsection{UTKFace}

\begin{figure*}[!tb]
\vskip 0.2in
\begin{center}
\centerline{\includegraphics[width=\textwidth]{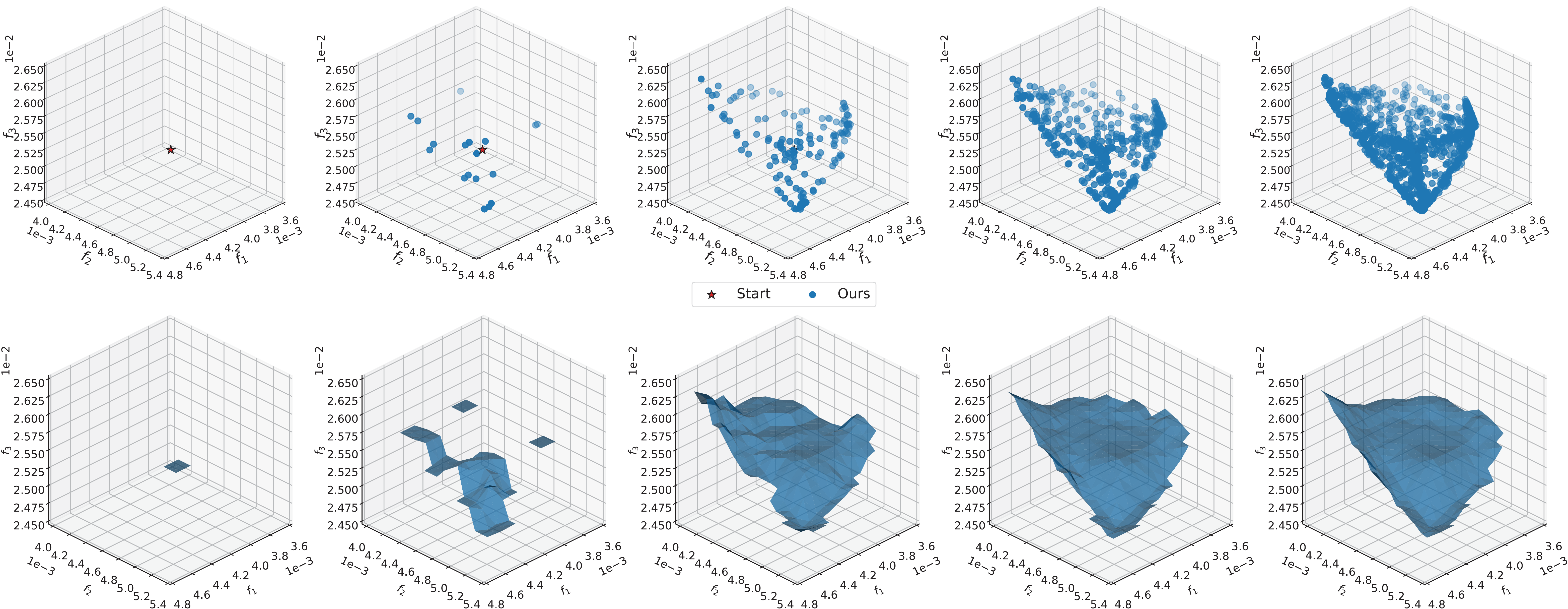}}
\caption{Continuous parametrization on UTKFace. The setup is identical to Figure \ref{fig:sup:uci_continuous} except that one Pareto seed obtained from ParetoMTL was used.}
\label{fig:sup:face_continuous}
\end{center}
\vskip -0.2in
\end{figure*}

Figure \ref{fig:sup:face_continuous} shows the continuous parametrization results on UTKFace. The setup and visualization is the same as in UCI Census-Income except that the continuous Pareto front was reconstructed from only 1 Pareto optimal seed. Therefore, a single color was used for all samples.

\section{Ablation Study}

\begin{figure*}[htb]
\vskip 0.2in
\begin{center}
\centerline{\includegraphics[width=\textwidth]{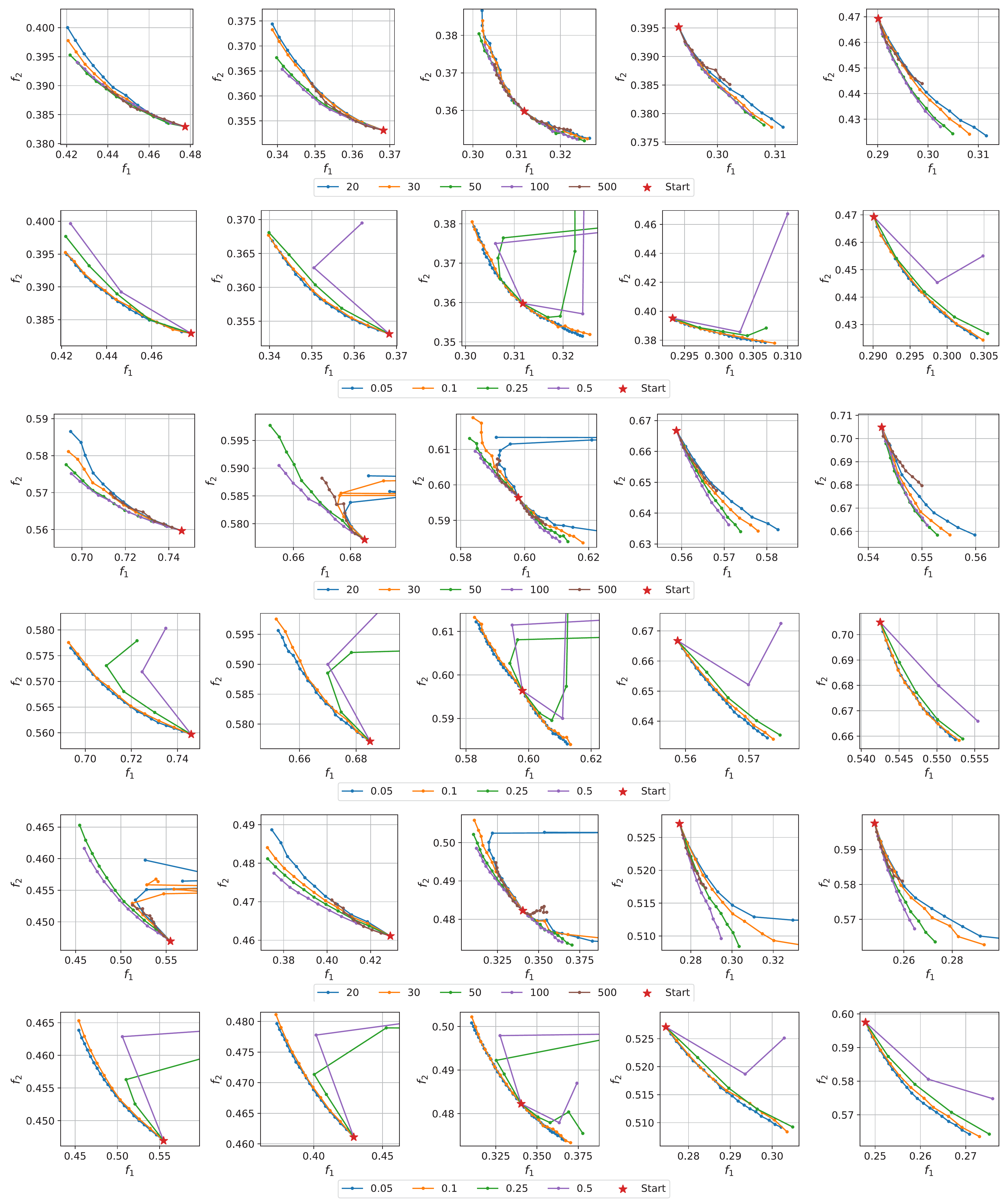}}
\caption{Ablation study on the maximum number of MINRES iterations $k$ and the step size $s$ on MultiMNIST (top two rows), MultiFashion (middle two rows), and MultiFashionMNIST (bottom two rows). We repeated the experiments from different Pareto optimal solutions returned by ParetoMTL (left to right).}
\label{fig:sup:ab_test}
\end{center}
\vskip -0.2in
\end{figure*}

Finally, we present more results of ablation study on MultiMNIST and its two variants in Figure \ref{fig:sup:ab_test}. For each dataset, we ran ParetoMTL to generate 5 Pareto optimal solutions that are evenly distributed on the Pareto front. From each solution, we conducted the ablation study on hyperparameters $k$ and $s$ as described in the main paper and produced one column of Figure \ref{fig:sup:ab_test}. It can be seen that our claims in the main paper on the influence of $k$ and $s$ are consistently observed across these 5 solutions with various trade-offs on these datasets.

\bibliography{main}
\bibliographystyle{icml2020}